\documentclass[journal]{IEEEtran}
\ifCLASSINFOpdf
\else
\fi

\usepackage{cite}
\usepackage{graphicx}
\usepackage{algorithmic}
\usepackage{algorithm}
\usepackage{amsmath}
\usepackage{amssymb}
\usepackage{MnSymbol}%
\usepackage{booktabs}
\usepackage{color}
\usepackage{threeparttable}
\usepackage{array}

\AtBeginDocument{%
	\providecommand\BibTeX{{%
			\normalfont B\kern-0.5em{\scshape i\kern-0.25em b}\kern-0.8em\TeX}}}

\usepackage{multirow}

\usepackage{enumitem}

\usepackage[dvipsnames]{xcolor}
\definecolor{redcolor}{rgb}{0.761, 0.328, 0.263}
\definecolor{bluecolor}{rgb}{0.192, 0.545, 0.886}

\usepackage{tikz}

\graphicspath{{figure/}}

\begin{document}
\bstctlcite{IEEEexample:BSTcontrol}
\title{Decoupled Doubly Contrastive Learning for Cross Domain Facial Action Unit Detection}

\author{Yong Li, Menglin Liu, Zhen Cui, Yi Ding, Yuan Zong,  Wenming Zheng,  Shiguang Shan ~\IEEEmembership{Fellow,~IEEE} and Cuntai Guan ~\IEEEmembership{Fellow,~IEEE}
	
	\IEEEcompsocitemizethanks{ \IEEEcompsocthanksitem 
		Corresponding author: Cuntai Guan, Shiguang Shan
	\IEEEcompsocthanksitem
	Yong Li is with the School of Computer Science and Engineering and the Key Laboratory of New Generation Artificial Intelligence Technology and Its Interdisciplinary Applications, Southeast University, Nanjing 210096, China. Email: mysee1989@gmail.com. 
	\IEEEcompsocthanksitem Menglin Liu is with the Key Laboratory of Intelligent Perception and Systems for High-Dimensional Information, Ministry of Education, School of Computer Science and Engineering, Nanjing University of Science and Technology, Nanjing, 210094, China. Email: 121106010799@njust.edu.cn.
	\IEEEcompsocthanksitem Zhen Cui is with the School of Artificial Intelligence, Beijing Normal University, Beijing, 100875, China. Email: zhen.cui@bnu.edu.cn. 
	\IEEEcompsocthanksitem  Shiguang Shan is with the State Key Laboratory of AI Safety, Institute of Computing Technology, Chinese Academy of Sciences, Beijing 100190, China, and with the University of Chinese Academy of Sciences, Beijing 100049, China. E-mail: sgshan@ict.ac.cn.
	\IEEEcompsocthanksitem Cuntai Guan and Yi Ding are with the School of Computer Science and Engineering, Nanyang Technological University, 50 Nanyang Avenue, Singapore, 639798. E-mail: {(ctguan, ding.yi)}@ntu.edu.sg.
	\IEEEcompsocthanksitem Wenming Zheng and Yuan Zong are with the Key Laboratory of Child Development and Learning Science of Ministry of Education, School of Biological Science and Medical Engineering, Nanjing 210096, China. E-mail: {(wenming\_zheng, xhzongyuan)}@seu.edu.cn.
	}
}

\markboth{Journal of \LaTeX\ Class Files,~Vol.~14, No.~8, August~2015}%
{Shell \MakeLowercase{\textit{et al.}}: Bare Demo of IEEEtran.cls for IEEE Journals}

\maketitle

\begin{abstract}
Despite the impressive performance of current vision-based facial action unit (AU) detection approaches, they are heavily susceptible to the variations across different domains and the cross-domain AU detection methods are under-explored.
In response to this challenge, we propose a decoupled doubly contrastive adaptation (D$^2$CA) approach to learn a purified AU representation that is semantically aligned for the source and target domains.
Specifically, we decompose latent representations into AU-relevant and AU-irrelevant components, with the objective of exclusively facilitating adaptation within the AU-relevant subspace. 
To achieve the feature decoupling, D$^2$CA is trained to disentangle AU and domain factors by assessing the quality of synthesized faces in cross-domain scenarios when either AU or domain attributes are modified.
To further strengthen feature decoupling, particularly in scenarios with limited AU data diversity, D$^2$CA employs a doubly contrastive learning mechanism comprising image and feature-level contrastive learning to ensure the quality of synthesized faces and mitigate feature ambiguities. This new framework leads to an automatically learned, dedicated separation of AU-relevant and domain-relevant factors, and it enables intuitive, scale-specific control of the cross-domain facial image synthesis.
Extensive experiments demonstrate the efficacy of D$^2$CA in successfully decoupling AU and domain factors, yielding visually pleasing cross-domain synthesized facial images. Meanwhile, D$^2$CA consistently outperforms state-of-the-art cross-domain AU detection approaches, achieving an average F1 score improvement of 6\%-14\% across various cross-domain scenarios.

\end{abstract}

\IEEEpeerreviewmaketitle

\section{Introduction}
\label{sec:introduction}

\begin{figure}[h!]
	\centering
	\includegraphics[width=1.0\linewidth]{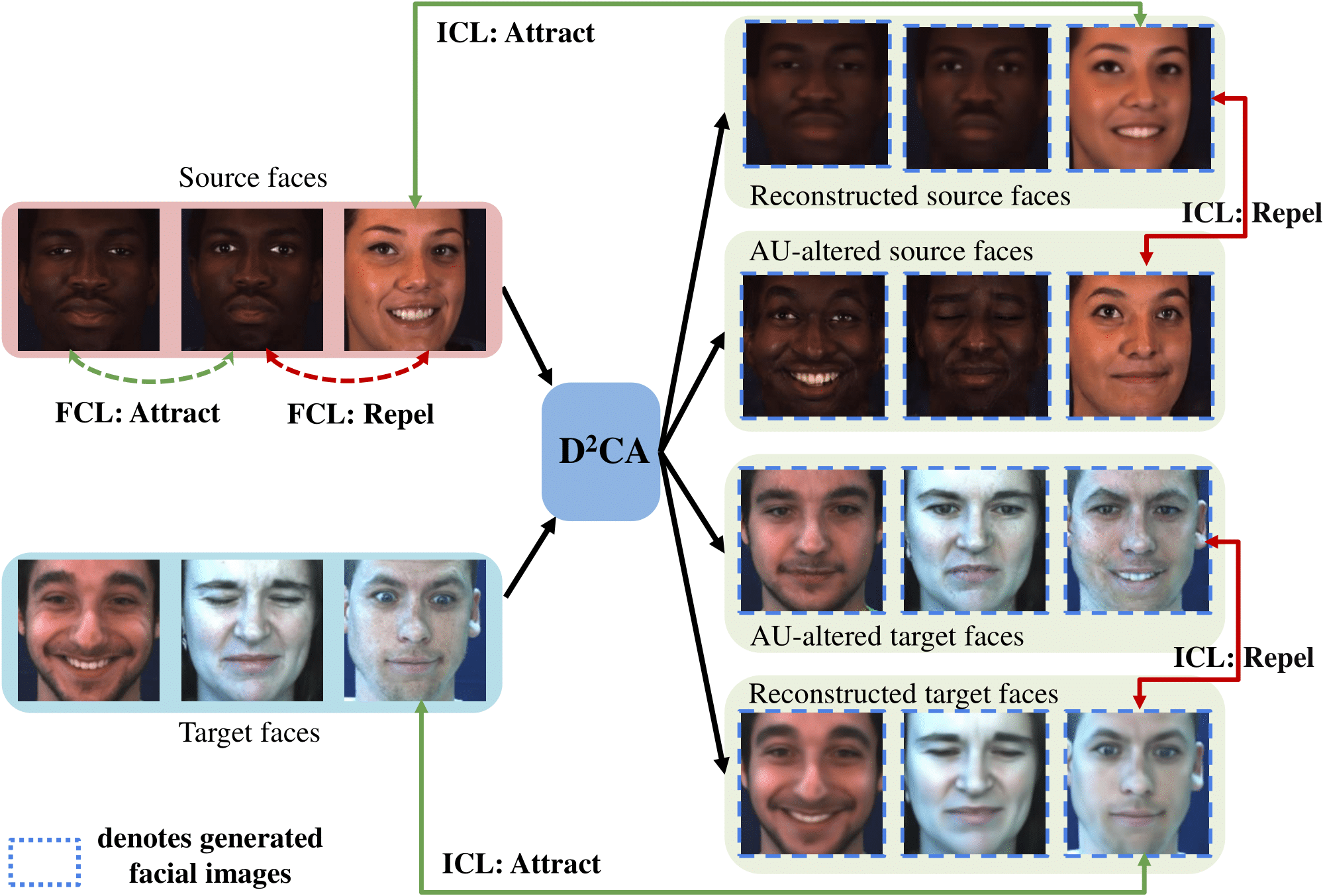}
	\caption{
		Main idea of our proposed Decoupled Doubly Contrastive Adaptation (D$^2$CA) method.
		D$^2$CA learns the decoupled AU-relevant features via encoding the AU-/domain-relevant features and conducting cross-domain face image generation.
		\textit{To consolidate the feature decoupling and adaptation}, D$^2$CA utilizes the doubly contrastive learning (CL) paradigm that consists of image (ICL) and feature (FCL) level contrastive learning to ensure the quality of the synthesized face images.
	}
	\label{fig:fig1_mainidea}
\end{figure}

Facial actions encompass a wide spectrum of nuanced meanings, encapsulating aspects such as an individual's intentions, emotions, attitudes, and their mental as well as physical states. The most comprehensive methodology for encoding and annotating these facial actions is the anatomically-grounded Facial Action Coding System (FACS)~\cite{friesen1978facial}. In FACS, facial action units (AUs) are delineated based on observable changes in facial appearance, encompassing alterations in geometric shape and textures resulting from facial muscle movements. The automated detection of AUs has garnered significant attention within the research community, given its abundant potential applications, including affect analysis and mental health assessment~\cite{tian2001recognizing, mcduff2010affect}.

In recent years, deep neural networks trained on extensive datasets have emerged as the prevailing paradigm for facial action unit (AU) detection~\cite{sun2021emotion, jacob2021facial, yang2023toward, li2022learning}. However, the current AU detection models exhibit significant susceptibility to domain variations, and there remains a dearth of comprehensive exploration in the realm of cross-domain AU detection methods~\cite{ghosh2015multi, ertugrul2020crossing}. Ertugrul \textit{et al.}~\cite{ertugrul2020crossing} performed comprehensive cross-domain AU detection experiments on several representative FACS datasets. The findings indicate that both deep and shallow approaches display limited cross-domain generalizability, suggesting it is deeply essential for enhanced adaptability across domains. 
Moreover, the process of labeling AUs is inherently challenging and prone to errors, given that AUs induce subtle facial deformations across various facial regions with varying intensities~\cite{ertugrul2020crossing, zhao2018learning}. To mitigate the need for manual AU annotations in the face of potential domain variations, there is a growing demand for models capable of adapting to new environments without the requirement of labeled data. Towards omitting manual AU annotations under all the possible domain variations, models that can adapt to new environments without labeled data are highly desirable.

\begin{figure*}[h!]
	\centering
	\includegraphics[width=1.0\linewidth]{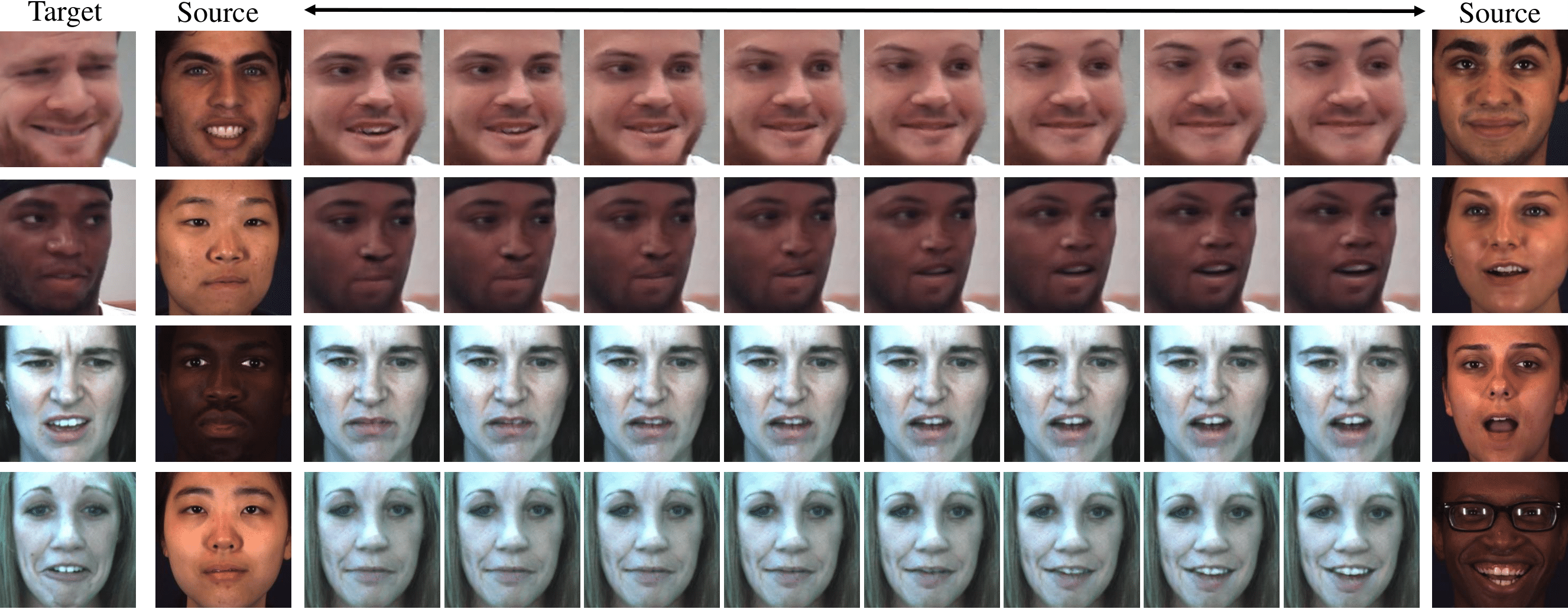}
	\caption{
		Results of domain-preserving AUs interpolation between two source faces.
		\textbf{Left column:} target face. \textbf{Second and Last column:} source faces.
		The synthesized target face indicates the AU-irrelevant representations have been well disentangled from the domain-relevant parts. Zoom in for details.
	}
	\label{fig:feature_interpolation}
\end{figure*}

The fundamental challenge in crafting an adaptable Facial Action Unit (AU) representation lies in the imperative of rendering it immune to domain variations. These variations encompass factors such as individual differences (e.g., gender, ethnicity, age), camera orientation (pose), illumination, video resolution, and more~\cite{ertugrul2019cross}. Inspired by these prior knowledge, our objective is to segregate the AU feature from the AU-irrelevant (domain-relevant) components and ensure that the adaptation process exclusively operates within the decoupled AU-relevant subspace.

Recent efforts have explored feature decoupling in contexts like person re-identification~\cite{zou2020joint, chen2021joint} or general image detection/segmentation~\cite{lee2021dranet} within the framework of unsupervised domain adaptation (UDA). However, it's important to note that (1) AU representation primarily captures fine-grained facial deformations, and (2) the desired AU-relevant representation may not conform to the classical style-content separation paradigm. Within the classical style-content separation framework~\cite{lee2021dranet}, image content is typically defined by high-level features or structures, such as the shape, layout, and identity of objects. Conversely, style pertains to lower-level features, including textures, colors, contrast, or lighting conditions. Thus, AUs should be regarded as part of the image content, as they represent dynamic changes in facial structure rather than static stylistic attributes.
Moreover, AU datasets typically consist of a limited number of individuals~\cite{zhao2018learning, sun2021emotion}, resulting in insufficient diversity within the training data. This limitation further complicates the task of cross-domain AU feature decoupling and adaptation.

To address these challenges, we introduce a novel approach termed ``Decoupled Doubly Contrastive Adaptation'' (D$^2$CA), which is designed to facilitate the learning of a decoupled and adaptable Facial Action Unit (AU) representation. As illustrated in Fig.~\ref{fig:fig1_mainidea}, given a pair of input images randomly sampled from source and target domains, we perform a decomposition of these images into two distinct spaces: AU-relevant (domain-irrelevant) and AU-irrelevant (domain-relevant). This decomposition is preliminarily achieved through a combination of a shared AU encoder and private domain encoders.

In order to accomplish feature decoupling, we integrate a cross-domain face synthesis mechanism utilizing AU and domain features from both source and target domains. As depicted in Fig.~\ref{fig:fig1_mainidea}, a synthesized AU-altered source face combines the AU characteristics from the target domain with the domain attributes from the source, and vice versa. Since real face images where only AU or domain is swapped are not readily available, we introduce a Cyclical Feature Alignment (CFA) mechanism that predicts the decoupled features from the AU-altered faces and subsequently predicts them in a self-supervised manner. Essentially, the quality of the synthesized AU-altered source or target faces serves as an indicator of how effectively the AU-relevant representation is decoupled and, consequently, how adaptable the feature becomes.


Nonetheless, when generating a cross-domain synthesized face that integrates AU and domain features from distinct domains, there is a risk of failure owing to the limited information available in the latent feature spaces.
To strengthen the feature decoupling, especially in scenarios characterized by insufficient diversity in AU data, we introduce a doubly contrastive learning strategy. This strategy encompasses both image-level Contrastive Learning (CL) and feature-level CL, with the objective of alleviating feature ambiguity and ensuring the high quality of the synthesized facial images.

\begin{figure*}[htb]
	\centering
	\includegraphics[width=1.0\linewidth]{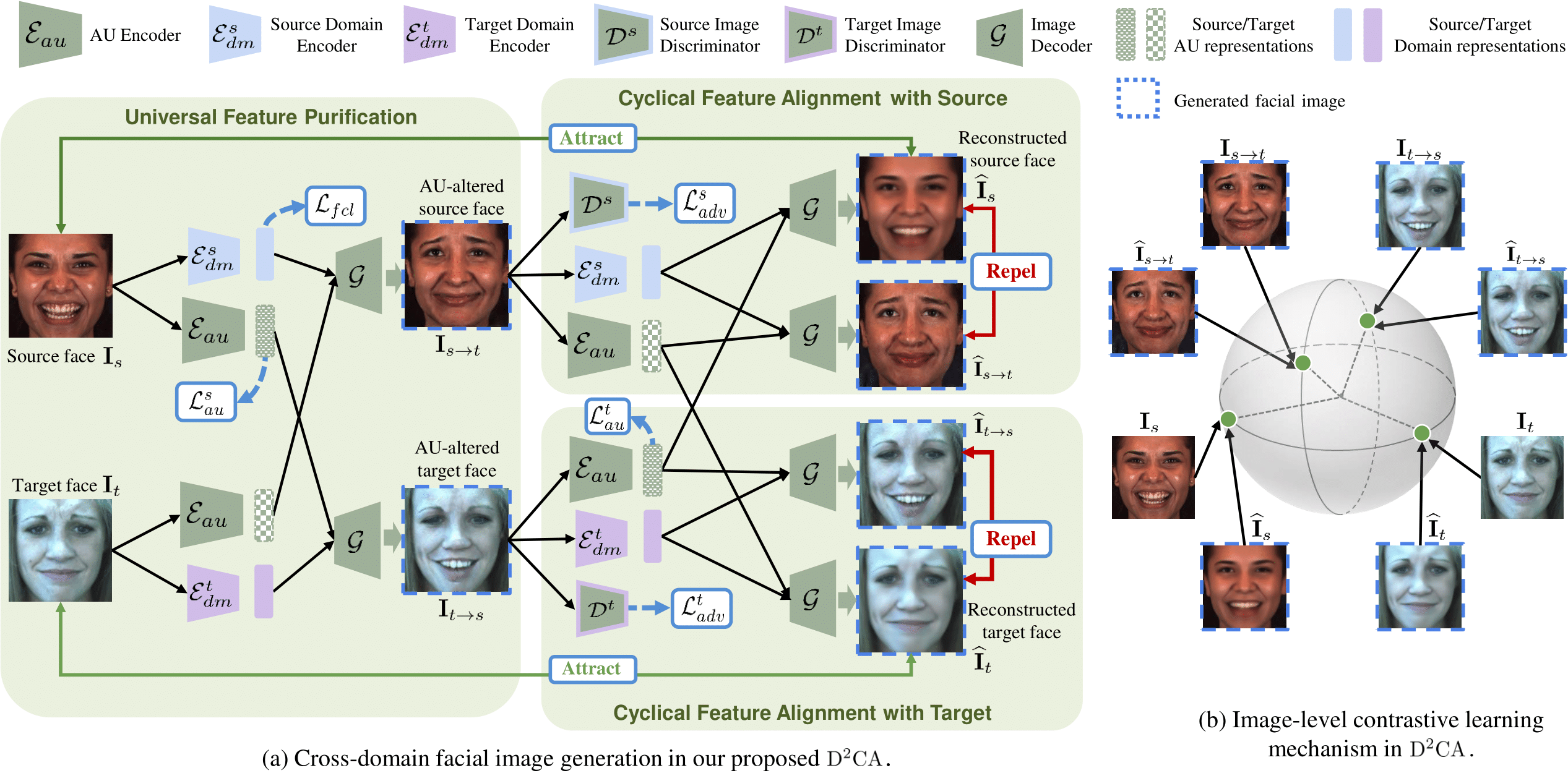}
	\caption{
		(a) shows the framework of D$^2$CA. Given a randomly sampled image pair, in \textit{Universal Feature Decoupling}, D$^2$CA encodes the AU-relevant (domain-irrelevant) and domain-relevant (AU-irrelevant) features via the shared AU encoder and exclusive domain encoders. Subsequently, the AU features obtained from different sources are exchanged to generate AU-altered facial images.
		In \textit{Cyclical Feature Alignment with source/target}, D$^2$CA utilizes the AU-altered images as input to anticipate the AU/domain features necessary for facial image regeneration and reconstruction. The primary objective here is to establish semantic alignment of AU features across different domains. 
		Sub-figure (b) highlights the application of image-level contrastive learning, where a facial image and its reconstructed counterpart are considered as positive samples, while others are regarded as negative samples.} 
	\label{fig:dca_framework}
\end{figure*}

As illustrated in Fig.~\ref{fig:fig1_mainidea},  
ICL leverages the input source face, the reconstructed source face, and the AU-altered source face as anchor, positive, and negative samples, respectively. ICL's primary objective is to ensure that the generated AU-altered source face is positioned far away from the input source face within the coupled representation space. This helps prevent the undesired mode collapse phenomenon in the synthesized faces. Concurrently, the feature-level CL enforces domain proximity associated with same identities and domain distance among different identities. This regulation is crucial to prevent the domain-relevant representation from degenerating into meaningless vectors, thereby preserving the encoding of domain-relevant factors. The process of cross-domain feature decoupling and adaptation is effectively consolidated through the doubly contrastive attraction and repulsion mechanism.
Through the innovative doubly contrastive learning paradigm, our proposed D$^2$CA effectively ensures the quality and fidelity of the cross-domain AU-altered facial images. 
In addition to its remarkable capability for generating visually appealing cross-domain facial images, D$^2$CA also enhances data diversity by combining limited existing identities with various Action Units (AUs) across different domains. Our extensive experimental results affirm that D$^2$CA represents an effective approach for acquiring semantically aligned, yet cross-domain, discriminative AU representations.
As validated in each row of Fig.~\ref{fig:feature_interpolation}, we generate a series of AU-altered target faces by interpolating between the AU representations of two source faces. The resulting interpolated target faces, as demonstrated in Fig.~\ref{fig:feature_interpolation}, exhibit smoothly transitioning AUs while retaining the unaltered domain characteristics.

The contributions of this work can be summarized as:
\begin{itemize}
	\item We propose a decoupled doubly contrastive adaptation (D$^2$CA) framework for unsupervised cross-domain AU detection.  This framework incorporates image and feature-level contrastive learning (CL) to ensure feature decoupling and cross-domain alignment.
	\item We introduce a cross-domain Cyclical Feature Alignment (CFA) mechanism to establish the foundation for effective feature decoupling and novel face regeneration. The CFA is seamlessly integrated with Doubly Contrastive Learning (DCL), which facilitates the synthesis of visually appealing cross-domain facial images. By incorporating both CFA and DCL, D²CA enhances data diversity by combining existing identities with various AUs across different domains.
	\item Extensive experiments validate the two contrastive learning mechanisms functions  collaboratively to ensure cross-domain feature alignment. Additionally, both quantitative and qualitative comparisons demonstrate the method’s ability to successfully decouple AU and domain factors, even in scenarios with insufficient diversity in the training AU data.
\end{itemize}

\section{Related Work}

\subsection{Unsupervised Domain Adaptation}
Typical UDA methods can be roughly classified as \textit{input-level} and or \textit{feature-level} adaptation. The former generates new images with the content of source and the
style of target images via adversarial training~\cite{bousmalis2017unsupervised, hoffman2018cycada}. The models are subsequently adapted by training with style-transferred images. 
UDA at the feature level often minimizes the difference between the feature space statistics of the source and target. 
Ganin \textit{et al.}~\cite{ganin2015unsupervised} proposed the classical Domain-Adversarial Neural Network. It consists of a gradient reversal layer that ensures the source and target feature distributions are aligned. Saito \textit{et al.}~\cite{saito2018maximum} introduced to align the source and target distributions by maximizing the discrepancy between two classifiers’ outputs. 
These prior efforts focus on matching global source and target data distributions to learn domain-invariant representations. However, the learned representations do not imply a fine-grained class-to-class alignment. 
Ye \textit{et al.} ~\cite{ye2020light} combined feature and input-level UDA to adjust target domain images, ensuring that their representations closely resemble those from the source domain. 
Recently, the search for feature decoupling-based UDA approaches has grown in interest~\cite{zou2020joint, chen2021joint,lee2021dranet, wu2021vector, gonzalez2018image}. Lee \textit{et al.}~\cite{lee2021dranet} proposed to disentangle image representations and transfer the visual attributes in a latent space for UDA.
Zou \textit{et al.}~\cite{zou2020joint} introduced a joint learning framework that separates id-relevant/-irrelevant features for person re-identification.
DRL~\cite{li2021unsupervised} utilizes implicit autoencoders that combine adversarial training with conditional generative models to achieve alignment of representations between the source and target domains.
DiCyR~\cite{bertoin2022disentanglement} employs a strategy of disentangling input information into two distinct components: task-related features, which are crucial for the learning task and style-related features, which are deemed task-irrelevant. While classical style-content separation methods are effective for general domain adaptation, they may not provide the level of fine-grained feature decoupling required for AU detection, where capturing small variations is critical.

For AU detection, a lot of studies have been developed and achieved significant progress~\cite{yang2023toward, jacob2021facial, li2018occlusion}. To mitigate the  data scarcity issue, several studies~\cite{li2019self, koepke2018self} have employed the self-supervised learning approach, wherein a target image is modified to resemble a novel image with a pose or expression akin to that of a source image. Notably, these methods necessitates the use of paired input images featuring identical identities. However, it is commonly observed that access to a restricted source image and an unrestricted target image sharing the same identity is often restricted. 
For cross-domain AU detection, Shao \textit{et al.}~\cite{shao2021unconstrained} proposed to decompose the source and target images into landmark-free/-related features and used feature swapping for UDA. Yin \textit{et al.}~\cite{yin2021self} proposed a self-supervised patch localization framework for UDA AU detection.
In contrast to the conventional paradigm based on style-content separation, this paper introduces a novel approach that employs a doubly contrastive learning strategy. This strategy is designed to enhance feature decoupling, particularly in situations characterized by a lack of diversity in AU data. Besides, our proposed D$^2$CA operates independently of facial landmark information, with the quality of the generated novel images serving as a direct indicator of the efficacy of feature purification.

\subsection{Contrastive Learning}
Recently, contrastive learning (CL) based self-supervised representation learning methodologies have demonstrated considerable potential~\cite{wu2018unsupervised,chen2020simple,he2020momentum,chen2021exploring,jing2020self, zhang2022dual, caron2020unsupervised, zbontar2021barlow, grill2020bootstrap, xiao2023degrade, jiang2020dual}. The fundamental objective of CL is to establish an embedding space wherein similar sample pairs remain closely clustered, while dissimilar pairs are distinctly separated. Typically, self-supervised learning methods generate internally defined pseudo labels to guide the acquisition of representations that prove applicable in subsequent tasks. 
These methods commonly center around an instance discrimination task, wherein a query is identified as matching a key if both correspond to the same image under various transformations. Building upon this paradigm, Chen \textit{et al.} \cite{chen2020simple} introduced the SimCLR framework, where a base encoder network and a projection head are trained to enhance agreement through a contrastive loss.

Similarly, He \textit{et al.} \cite{he2020momentum} proposed Momentum Contrast (MoCo). Typically, MoCo incorporates a momentum network that manages a queue housing a substantial number of negative samples, thus enabling efficient contrastive learning. Subsequently, Caron \textit{et al.}~\cite{caron2020unsupervised} introduced a more memory-efficient approach, known as Swapping Assignments between Views (SwAV), which concurrently clusters data while enforcing consistency between cluster assignments derived for different augmentations of the same image, as opposed to directly comparing features as in contrastive learning.
To circumvent trivial constant solutions in contrastive learning, Zbontar \textit{et al.}~\cite{zbontar2021barlow} proposed an objective function that naturally averts representation collapse by evaluating the cross-correlation matrix between the outputs of two identical networks fed with distorted versions of a sample, aiming to make it as similar to the identity matrix as possible.
Diverging from the aforementioned methods, Grill \textit{et al.}~\cite{grill2020bootstrap} introduced Bootstrap Your Own Latent (BYOL), directly bootstrapping representations without relying on negative pairs or pseudo-labels. Furthermore, Chen \textit{et al.} \cite{chen2021exploring} pursued the acquisition of general representations without relying on negative sample pairs, implementing a stop-gradient operation to prevent undesired collapsing solutions. Xiao \textit{et al.}~\cite{xiao2023degrade} presented a self-supervised degradation-guided adaptive network for blind remote sensing image super-resolution. It addresses real-world degradation challenges by using contrastive learning to derive robust representations of various degradations.
Drawing upon the contrastive principle, this study introduces the doubly contrastive learning (DCL) strategy, ensuring the effectiveness of cross-domain synthesized facial images without relying on specific data augmentation techniques. The proposed DCL strategy serves as a foundation for  effective feature decoupling and cross-domain AU detection adaptation.

\section{The Proposed Method}

The framework of D$^2$CA is illustrated in Fig.~\ref{fig:dca_framework} (a).  As can be seen, D$^2$CA mainly consists of three core components: (1) \textbf{Universal Feature Purification}; (2) \textbf{Cyclical Feature Alignment with Source}; (3) \textbf{Cyclical Feature Alignment with Target}.

During the \textbf{Universal Feature Purification} phase, when presented with a pair of images randomly selected from the source and target domains, D$^2$CA is directed to generate AU-altered images of the source and target faces through feature interchanging in the latent space. A comprehensive explanation of the feature purification process is presented in Sec.~\ref{sec:feature_decoupling}. 

\begin{figure}[htb]
	\centering
	\includegraphics[width=1.0\linewidth]{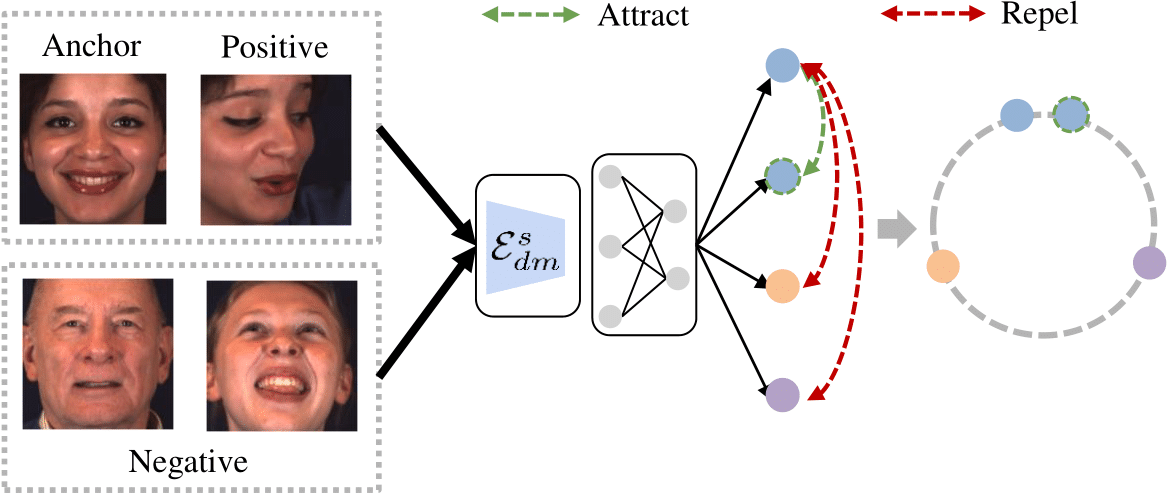}
	\caption{
		Illustration of feature-level contrastive learning (FCL). Domain features from the same identity are pushed close.
	}
	\label{fig:feature_cl}
\end{figure}

Subsequently, In the \textbf{Cyclical Feature Alignment with Source} and \textbf{Cyclical Feature Alignment with Target} stages, D$^2$CA predicts the purified AU and domain features from the AU-altered facial images for the purpose of face re-generation and reconstruction. Specifically, D$^2$CA aims to generate an AU-altered target facial image with AU attributes derived from the source face, enabling AU supervision within the target domain. This approach establishes semantic alignments between the purified source and target AU features, facilitating cross-domain adaptation. Further elucidation of these procedures can be found in Sec.~\ref{sec:sr_au_domain}.
Considering the absence of actual facial images wherein only the AU or domain is swapped, we introduce a doubly contrastive learning (DCL) paradigm to consolidate the feature decoupling and cross-domain adaptation. DCL comprises image-level CL (ICL, shown in Fig.~\ref{fig:dca_framework} (b)) and feature-level CL (FCL, shown in Fig.~\ref{fig:feature_cl}), aiming to guarantee the quality of the AU-altered source/target facial images (Sec.~\ref{sec:dcl}).

Below, we present details of the three parts of D$^2$CA.
\subsection{Universal Feature Purification}
\label{sec:feature_decoupling}

As illustrated in Fig.~\ref{fig:dca_framework},  D$^2$CA leverages a shared AU encoder $\mathcal{E}_{au}$ and two private domain encoders, $\mathcal{E}_{dm}^s$, $\mathcal{E}_{dm}^t$, to explicitly forecast the refined AU representations $\mathcal{E}_{au}(\mathbf{I}_{s})$ and $\mathcal{E}_{au}(\mathbf{I}_{t})$, as well as  domain representations $\mathcal{E}_{dm}^s(\mathbf{I}_{s})$ and $\mathcal{E}_{dm}^t(\mathbf{I}_{t})$ from a given pair of randomly sampled source and target facial images ($\mathbf{I}_s$, $\mathbf{I}_t$).

Considering that the anticipated features are expected to encode the distinct AU and domain attributes separately, we establish a soft orthogonality constraint to diminish the redundancy of information between these two feature components:
\begin{equation}
	\mathcal{L}_{\text{ort}} = \mathbb{E}[\cos(\mathcal{E}_{au}(\mathbf{I}_{s}), \mathcal{E}_{dm}^s(\mathbf{I}_{s}))] + \mathbb{E}[\cos(\mathcal{E}_{au}(\mathbf{I}_{t}), \mathcal{E}_{dm}^t(\mathbf{I}_{t}))],
	\label{equ:orthogonality}
\end{equation}
where $\cos(\cdot,\cdot)$ means the cosine similarity between two
feature vectors. Following the encoding of features for both domains, we incorporate a cross-domain facial image generation paradigm to facilitate feature decoupling. Specifically, we interchange the AU and domain features between different domains to create novel facial images. The quality of the generated cross-domain AU-altered facial images indicate the effectiveness of the feature purification process and the adaptability of the AU feature.

As illustrated in Fig.~\ref{fig:dca_framework}, D$^2$CA generates an AU-altered source facial image $\mathbf{I}_{s \rightarrow t}$ using the AU feature $\mathcal{E}_{au}(\mathbf{I}_{t})$ from the target facial image $\mathbf{I}_{t}$ and the domain feature $\mathcal{E}_{dm}^s(\mathbf{I}_{s})$ from the source $\mathbf{I}_{s}$. Symmetrically, we synthesize an AU-altered target facial image via concatenating $\mathcal{E}_{au}(\mathbf{I}_{s})$ and  $\mathcal{E}_{dm}^t(\mathbf{I}_{t})$ and feed them into the shared image decoder $\mathcal{G}$. Formally,
\begin{equation}
	\mathbf{I}_{s \rightarrow t} = \mathcal{G}([\mathcal{E}_{au}(\mathbf{I}_{t}), \mathcal{E}_{dm}^s(\mathbf{I}_{s})]), 
\end{equation}
\begin{equation}
	\mathbf{I}_{t \rightarrow s} = \mathcal{G}([\mathcal{E}_{au}(\mathbf{I}_{s}), \mathcal{E}_{dm}^t(\mathbf{I}_{t})]), 
\end{equation}
where the notation $[.]$ means feature concatenation. The generated AU-altered target facial image $\mathbf{I}_{t \rightarrow s}$ mimics the AU attributes derived from the source face and domain from the target.  Given the availability of AU annotations for the source facial images, the sufficient generation quality of $\mathbf{I}_{t \rightarrow s}$ would allow us to undertake AU supervision within the target domain accordingly. 

For the AU representations from the source face images, we exploit the multi-label sigmoid cross-entropy loss for supervised AU detection. Formally,
\begin{equation}
	\mathcal{L}_{au}^s = \mathbb{E}[- \sum^{\text{A}}_{a} z_a \log\hat{z}_{a} + (1 - z_{a}) \log (1-\hat{z}_{a})],
\end{equation}
where $\text{A}$ is the number of AUs. $z_a$ denotes the $a$-th ground truth AU label for the input source face image. $\hat{z}_a$ means the predicted AU score. 

\begin{table}[htb]
	 \centering
	\caption{F1-scores of domain adaptation from BP4D to GFT. \textbf{Bold} means the best.}
	\label{tab:bp4d_gft}
	\setlength\tabcolsep{2.0pt}
	\begin{center}
			\begin{tabular}{c|c|c|c|c|c|c|c|c|c|c|c}
				\toprule
				AU & 1 & 2 & 4 & 6 & 10 & 12 & 14 & 15 & 23 & 24 & AVE\\
				\toprule
				\multicolumn{12}{c}{Source: \textbf{BP4D}. Target: \textbf{GFT}. } \\
				\midrule
				\textit{Source}~\cite{yin2021self} & \textit{3.9} & \textit{7.6} & \textit{2.7} & \textit{44.8} & \textit{46.3} & \textit{38.0} & \textit{6.2} & \textit{19.2} & \textit{28.6} & \textit{18.7} & \textit{21.6}\\
				\midrule
				DANN~\cite{ganin2015unsupervised} & 5.9 & 17.4 & 2.8 & 51.4 & 38.3 & 52.4 & 6.0 & 18.1 & 36.1 & 20.9 & 24.9\\
				
				MCD~\cite{saito2018maximum} & 4.6 & 16.5 & 8.9 & 41.7 & 40.5 & 58.5 & 6.1 & 21.8 & 36.3 & 22.7 & 25.8\\
				
				P-MCD~\cite{yin2021self} & 9.6 & \textbf{22.1} & 8.1 & 54.0 & 53.6 & 57.2 & \textbf{7.7} & 14.4 & 37.0 & 31.8 & 29.6\\
				AFS~\cite{zhuo2023adaptive} & 16.5 & 17.6 & 9.7 & 54.1 & 44.7 & 56.0 & 7.1 & 15.5 & 27.7 & 43.8 & 29.3 \\
				CAF~\cite{xie2023collaborative} & 15.3 & 20.3 & 4.6 & 62.5 & 52.9 & 62.5 & 8.2 & 24.8 & 39.2 & 44.0 & 33.4 \\
				\midrule
				TCAE ~\cite{li2019self} & \textbf{17.4} & 19.4 & 9.5 & 62.0 & 54.6 & 59.3 & 7.1 & 10.4 & 37.3 & 36.7 & 31.4\\
				DGNet++~\cite{zou2020joint} & 6.3 & 16.8 & 7.7 & 65.3 & 55.3 & 60.5 & 5.2 & 12.9 & 34.7 & 37.0 & 30.2\\
				
				DRANet~\cite{lee2021dranet} & 13.0 & 18.4 & 8.4 & 61.3 & 56.1 & 55.8 & 7.3 & 19.7 & 32.8 & 40.3 & 31.3\\
				
				GCL~\cite{chen2021joint} & 14.2 & 4.3 & 7.7 & 61.4 & 52.4 & 61.1 & 6.1 & 23.4 & 24.2 & 44.5 & 29.9\\
				
				\textbf{D$^2$CA (Ours)} & 14.3 & 20.4 & \textbf{10.1} & \textbf{66.8} & \textbf{57.0} & \textbf{65.1} & 7.6 & \textbf{24.9} & \textbf{41.4} & \textbf{45.7} & \textbf{35.3} \\
				\midrule
				\textit{Target}~\cite{yin2021self} & \textit{13.2} & \textit{27.3} & \textit{9.8} & \textit{69.8} & \textit{70.8} & \textit{77.5} & \textit{8.9} & \textit{28.9} & \textit{47.4} & \textit{45.3} & \textit{39.9} \\
				\bottomrule
				\toprule
				\multicolumn{12}{c}{Source: \textbf{GFT}. Target: \textbf{BP4D}. } \\
				\midrule
				\textit{Source}~\cite{yin2021self} & \textit{28.7} & \textit{29.2} & \textit{27.6} & \textit{53.5} & \textit{49.0} & \textit{65.5} & \textit{11.5} & \textit{29.4} & \textit{33.0} & \textit{36.3} & \textit{36.4}\\
				\midrule
				DANN~\cite{ganin2015unsupervised} & 12.7 & 17.3 & 10.3 & 53.5 & 50.8 & 61.7 & 5.9 & 22.4 &  32.1 & 36.0 & 30.3\\
				
				MCD~\cite{saito2018maximum} & 19.0 & 11.7 & 11.6 & 60.3 & 52.3 & 63.2 & 14.3 & 17.6 & 35.4 & 31.7 & 31.7\\
				
				P-MCD~\cite{yin2021self} & 19.2 & 11.6 & 18.5 & 61.5 & 50.3 & 68.0 & 14.0 & 19.4 & 35.1 & 39.5 & 33.7\\
				\midrule
				TCAE ~\cite{li2019self} & 26.3 & 30.7 & 20.6 & 57.6 & 52.8 & 65.6 & 10.7 & 28.2 & 24.0 & 37.1 & 35.4\\
				DGNet++~\cite{zou2020joint} & 27.3 & 29.4 & 38.4 & 54.8 & 51.1 & 53.8 & 20.1 & 34.6 & 24.2 & 33.9 & 36.8\\
				
				DRANet~\cite{lee2021dranet} & 29.0 & 32.0 & 19.8 & \textbf{63.5} & 52.0 & 69.7 & 13.3 & 33.4 & 33.4 & 41.4 & 38.8\\
				
				GCL~\cite{chen2021joint} & 30.8 & 28.8 & 40.1 & 52.5 & 46.2 & 57.3 & 16.0 & \textbf{36.6} & 32.0 & \textbf{45.6} & 38.6\\
				\textbf{D$^2$CA (Ours)} & \textbf{31.0} & \textbf{33.0} & \textbf{43.2} & 60.5 & \textbf{53.7} & \textbf{71.1} & \textbf{21.1} & 36.5 & \textbf{36.4} & 44.6 & \textbf{43.1}\\
				\midrule
				
				\textit{Target}~\cite{yin2021self} & \textit{56.6} & \textit{50.2} & \textit{36.0} & \textit{74.0} & \textit{82.2} & \textit{83.9} & \textit{52.5} & \textit{39.3} & \textit{38.6} & \textit{40.2} & \textit{55.4}\\
				\bottomrule
			\end{tabular}
	\end{center}
\end{table}

\begin{table}[htb]
	\caption{F1-scores of domain adaptation from BP4D+ to GFT. \textbf{Bold} means the best.}
	\label{tab:bp4d_plus_gft}
	\setlength\tabcolsep{2.0pt}
	\begin{center}
			\begin{tabular}{c|c|c|c|c|c|c|c|c|c|c|c}
				\toprule
				AU & 1 & 2 & 4 & 6 & 10 & 12 & 14 & 15 & 23 & 24 & AVE\\
				\toprule
				\multicolumn{12}{c}{Source: \textbf{BP4D+}. Target: \textbf{GFT}. } \\
				\midrule
				\textit{Source}~\cite{yin2021self} & \textit{4.5} & \textit{6.1} & \textit{2.9} & \textit{43.2} & \textit{41.2} & \textit{49.4} & \textit{6.3} & \textit{14.9} & \textit{42.2} & \textit{14.3} & \textit{22.5}\\
				
				\midrule
				DANN~\cite{ganin2015unsupervised} & 8.3 & 12.9 & 10.7 & 54.9 & 48.7 & 56.1 & 7.0 & 8.8 & 33.8 & 16.9 & 25.9\\
				MCD~\cite{saito2018maximum} & 5.1 & 15.8 & 9.5 & 55.5 & 48.2 & 61.0 & 6.6 & 14.9 & 31.0 & 25.8 & 27.3\\
				P-MCD~\cite{yin2021self} & 7.1 & 16.9 & 10.4 & 54.0 & 54.6 & 59.1 & 6.1 & 14.5 & 40.4 & 31.8 & 29.5\\
					AFS~\cite{zhuo2023adaptive} & 15.2 & 31.2 & 9.1 & 58.7 & 64.7 & 64.2 & 7.7 & 18.0 & 38.8 & 30.1 & 33.8 \\
				CAF~\cite{xie2023collaborative} & 14.9  & 21.5  & 10.2  & 63.2  & 68.0  & 68.4  & 5.2  & 24.2  & 44.0  & 26.5  & 34.6 \\
				\midrule
				TCAE ~\cite{li2019self} & 11.5 & 27.1 & 11.2 & \textbf{65.3} & 66.3 & 66.9 & 5.4 & 25.8 & 38.8 & 32.3 & 35.1\\
				DGNet++~\cite{zou2020joint} & 11.1 & 20.9 & 5.9 & 61.2 & 68.5 & 67.6 & 2.6 & 25.4 & 41.8 & 31.9 & 33.7\\
				DRANet~\cite{lee2021dranet} & 14.8 & 27.3 & \textbf{13.7} & 61.5 & 56.8 & 66.9 & 4.2 & 14.4 & 33.2 & 30.7 & 32.4\\
				GCL~\cite{chen2021joint} & 13.7 & 29.1 & 5.9 & 62.6 & 54.7 & 64.1 & 4.2 & 24.5 & 40.3 & 28.1 & 32.7\\
				\textbf{D$^2$CA (Ours)} & \textbf{15.7} & \textbf{37.4} & 7.5 & 63.7 & \textbf{68.8} & \textbf{68.7} & \textbf{8.2} & \textbf{28.0} & \textbf{45.9} & \textbf{33.7} & \textbf{37.8} \\
				\midrule
				\textit{Target}~\cite{yin2021self} & \textit{13.2} & \textit{27.3} & \textit{9.8} & \textit{69.8} & \textit{70.8} & \textit{77.5} & \textit{8.9} & \textit{28.9} & \textit{47.4} & \textit{45.3} & \textit{39.9}\\
			\bottomrule
			\toprule
			\multicolumn{12}{c}{Source: \textbf{GFT}. Target: \textbf{BP4D+}. } \\
			\midrule
			\textit{Source}~\cite{yin2021self} & \textit{14.1} & \textit{14.4} & \textit{13.7} & \textit{54.8} & \textit{66.9} & \textit{70.6} & \textit{6.4} & \textit{28.7} & \textit{32.7} & \textit{16.9} & \textit{31.9}\\
			
			\midrule
			DANN~\cite{ganin2015unsupervised} & 12.6 & 10.6 & 1.0 & 61.2 & 63.5 & 71.7 & 3.3 & 18.4 & 30.5 & 12.8 & 28.6\\
			MCD~\cite{saito2018maximum} & 16.5 & 12.2 & 10.3 & 68.2 & 68.9 & 75.8 & 8.9 & 21.5 & 32.8 & 12.6 & 32.8\\
			P-MCD~\cite{yin2021self} & 16.6 & 14.5 & 12.2 & \textbf{68.3} & 67.1 & 75.1 & 7.5 & 22.9 & 35.1 & 14.3 & 33.4\\
			\midrule
			TCAE ~\cite{li2019self} & 20.3 & 17.8 & 5.9 & 52.9 & \textbf{72.6} & 72.8 & 9.1 & 24.5 & 24.0 & 15.1 & 31.5\\
			DGNet++~\cite{zou2020joint} & 17.5 & 15.4 & 6.9 & 66.7 & 67.2 & 74.4 & 23.7 & 31.6 & 29.4 & 11.8 & 34.5\\
			DRANet~\cite{lee2021dranet} & 20.8 & 17.8 & 9.0 & 61.0 & 64.2 & 71.3 & 9.2 & 36.3 & \textbf{40.0} & 21.1 & 35.1\\
			GCL~\cite{chen2021joint} & 15.5 & 16.9 & 15.0 & 60.7 & 70.6 & 69.3 & 4.4 & 34.4 & 39.0 & 21.7 & 34.8\\
			
			\textbf{D$^2$CA (Ours)} & \textbf{22.8} & \textbf{19.3} & \textbf{18.7} & 59.8 & 72.0 & \textbf{77.5} & \textbf{26.7} & \textbf{36.7} & 39.0 & \textbf{22.7} & \textbf{39.5}\\
			\midrule
			
			\textit{Target}~\cite{yin2021self} & \textit{41.5} & \textit{30.6} & \textit{37.1} & \textit{84.6} & \textit{89.0} & \textit{88.9} & \textit{73.8} & \textit{42.2} & \textit{51.3} & \textit{23.5} & \textit{56.3}\\
			\bottomrule
			\end{tabular}
	\end{center}
\end{table}

\subsection{Cyclical Feature Alignment with Source and Target}
\label{sec:sr_au_domain}
Since we have no pixel or label level supervisions for the cross-domain generated AU-altered source facial image $\mathbf{I}_{s \rightarrow t}$ or AU-altered target facial image  $\mathbf{I}_{t \rightarrow s}$, we introduce a  Cyclical Feature Alignment (CFA) mechanism that predicts the decoupled representations from $\mathbf{I}_{s \rightarrow t}$ and $\mathbf{I}_{t \rightarrow s}$, then  undertake feature swapping in the latent space again.

As illustrated in Fig.~\ref{fig:dca_framework} (a), D$^2$CA contains two symmetric CFA. In CFA with source or target branch, we extract the AU and domain features of $\mathbf{I}_{s \rightarrow t}$ or $\mathbf{I}_{t \rightarrow s}$ using the shared AU encoder $\mathcal{E}_{au}$ and the private domain encoders, i.e., $\mathcal{E}_{dm}^s$ and $\mathcal{E}_{dm}^t$.
As the AU-altered source face $\mathbf{I}_{s \rightarrow t}$ and the source face $\mathbf{I}_{s}$ come from the same domain. Thus, their domain features should be similar. Meanwhile, $\mathbf{I}_{s \rightarrow t}$ and the target facial image $\mathbf{I}_{t}$ should illustrate same AUs. Thus, their AU features should be similar.
Formally, the AU and domain feature discrepancy for the AU-altered source facial image $\mathbf{I}_{s \rightarrow t}$ can be formulated as:
\small
\begin{equation}
	\label{eq:cyc_au_changed_source}
	\mathcal{L}^{s}_{rep} = \mathbb{E}[||\mathcal{E}_{au}(\mathbf{I}_{t}) - \mathcal{E}_{au}(\mathbf{I}_{s \rightarrow t}) ||^2 + ||\mathcal{E}_{dm}^{s}(\mathbf{I}_{s}) - \mathcal{E}_{dm}^{s}(\mathbf{I}_{s \rightarrow t})||^2].
\end{equation}
\normalsize

Similarly, the AU-changed target face $\mathbf{I}_{t \rightarrow s}$ should have similar AU features to the source face image $\mathbf{I}_{s}$, and have similar domain features with the target face image $\mathbf{I}_{t}$. The feature discrepancy for $\mathbf{I}_{t \rightarrow s}$ can be formulated as:
\small
\begin{equation}
	\label{eq:cyc_au_changed_target}
	\mathcal{L}^{t}_{rep} = \mathbb{E}[||\mathcal{E}_{au}(\mathbf{I}_{s}) - \mathcal{E}_{au}(\mathbf{I}_{t \rightarrow s}) ||^2 + ||\mathcal{E}_{dm}^{t}(\mathbf{I}_{t}) - \mathcal{E}_{dm}^{t}(\mathbf{I}_{t \rightarrow s})||^2].
\end{equation}
\normalsize

Besides the self-supervised constraints in the latent space, we exploit the pixel-level supervision between $\mathbf{I}_{s}$ and the reconstructed source facial image $\widehat{\mathbf{I}}_s$, as well as the discrepancy between $\mathbf{I}_{t}$ and $\widehat{\mathbf{I}}_t$. Formally,
\begin{equation}
	\label{eq:cyc_pixel}
	\begin{split}
		\mathcal{L}_{rec}^s = \mathbb{E}[||\mathbf{I}_{s} - \mathcal{G}([\mathcal{E}_{au}(\mathbf{I}_{t \rightarrow s}), \mathcal{E}_{dm}^{s}(\mathbf{I}_{s \rightarrow t})]) ||_1], \\
		\mathcal{L}_{rec}^t = \mathbb{E}[||\mathbf{I}_{t} - \mathcal{G}([\mathcal{E}_{au}(\mathbf{I}_{s \rightarrow t}), \mathcal{E}_{dm}^{t}(\mathbf{I}_{t \rightarrow s})]) ||_1],
	\end{split}
\end{equation}
where $\mathcal{L}_{rec}$ denotes the pixel-level discrepancy between the input face image and the reconstructed one.
$\widehat{\mathbf{I}}_s = \mathcal{G}([\mathcal{E}_{au}(\mathbf{I}_{t \rightarrow s}), \mathcal{E}_{dm}^{s}(\mathbf{I}_{s \rightarrow t})])$ means we synthesize $\widehat{\mathbf{I}}_s$ using the AU features of  $\mathbf{I}_{t \rightarrow s}$ and domain features of $\mathbf{I}_{s \rightarrow t}$, and vice versa for substituting $\widehat{\mathbf{I}}_t$ in Equ.~\ref{eq:cyc_pixel}.

In addition, we employ adversarial loss to align the distributions between the AU-changed facial images and the vanilla input source/target images. Formally,
\begin{equation}
	\label{eq:adversarial}
	\begin{split}
		\mathcal{L}_{adv}^s = \mathbb{E} [ \text{log} \mathcal{D}^s(\mathbf{I}_s) + \text{log} (1 - \mathcal{D}^s(\mathbf{I}_{s \rightarrow t})) ], \\
		\mathcal{L}_{adv}^t = \mathbb{E} [ \text{log} \mathcal{D}^t(\mathbf{I}_t) + \text{log} (1 - \mathcal{D}^t(\mathbf{I}_{t \rightarrow s})) ],
	\end{split}
\end{equation}
where $\mathcal{D}^s$ and $\mathcal{D}^t$ means two private PatchGAN-based image discriminator~\cite{isola2017image}, as shown in Fig.~\ref{fig:dca_framework} (a). These cross-domain generated facial images are adversarially trained to align with the distribution of real facial images.
Finally, we combine these  loss constraints to form the preliminary feature purifying constraints,
\begin{equation}
	\label{equ:decoupling}
	\mathcal{L}_{deco} =  \mathcal{L}_{au} + \gamma_1 \mathcal{L}_{rep} + \gamma_2 \mathcal{L}_{rec} + \gamma_3 (\mathcal{L}_{adv}^s + \mathcal{L}_{adv}^t) + \gamma_ 4\mathcal{L}_{ort},
\end{equation}
where $\mathcal{L}_{au}$ represents the combination of the AU detection loss of $\mathcal{L}_{au}^s$ and $\mathcal{L}_{au}^t$. The former and the latter respectively represent the supervision on the input source facial image $\mathbf{I}_s$ and the AU-altered target facial image $\mathbf{I}_{t \rightarrow s}$, as illustrated in Fig.~\ref{fig:dca_framework} (a). Besides, 
$\mathcal{L}_{rep}= \mathcal{L}_{rep}^s + \mathcal{L}_{rep}^t$ means the feature-level constraints in Equ.~\ref{eq:cyc_au_changed_source} and Equ.~\ref{eq:cyc_au_changed_target}.
Furthermore, $\mathcal{L}_{rec} = \mathcal{L}_{rec}^s + \mathcal{L}_{rec}^t$ denotes the linear combination of image-level constraints in Equ.~\ref{eq:cyc_pixel}. The hyperparameters $\gamma_1$, $\gamma_2$, $\gamma_3$ and $\gamma_4$ are the balance factors.

\begin{table}[htb]
	\caption{F1-scores of domain adaptation from BP4D to DISFA. \textbf{Bold} means the best.}
	\label{tab:bp4d_disfa}
	\setlength\tabcolsep{4.0pt}
	\begin{center}
		\scalebox{0.9}{
			\begin{tabular}{c|c|c|c|c|c|c}
				\toprule
				AU & 1 & 2 & 4 & 6 & 12 & AVE\\
			\toprule
			\multicolumn{7}{c}{Source: \textbf{BP4D}. Target: \textbf{DISFA}. } \\
			\midrule
				\textit{Source}~\cite{yin2021self} & \textit{11.5} & \textit{2.0} & \textit{14.7} & \textit{19.4} & \textit{43.2} & \textit{18.2} \\
				\midrule
				DANN~\cite{ganin2015unsupervised} & 10.1 & 8.1 & 17.1 & 33.3 & 56.8 & 25.1\\
				MCD~\cite{saito2018maximum} & 17.9 & 18.1 & 31.5 & 30.2 & 38.9 & 27.3\\
				P-MCD~\cite{yin2021self} & 34.3 & 16.6 & \textbf{52.1} & 33.5 & 50.4 & 37.4\\
				AFS~\cite{zhuo2023adaptive} & 25.6 & 11.6 & 26.7 & 33.9 & 59.0 & 31.4 \\
				CAF~\cite{xie2023collaborative} & 28.5 & 8.2 & 45.9 & 31.6 & 60.3 & 34.9 \\
				\midrule
				TCAE~\cite{li2019self} & 22.2 & 37.6 & 42.1 & 32.7 & 61.9 & 39.3\\
				DGNet++~\cite{zou2020joint} & 17.9 & 9.6 & 38.0 & 28.2 & 64.8 & 31.7\\
				DRANet~\cite{lee2021dranet} & 33.1 & 27.0 & 49.2 & 33.4 & 65.7 & 41.7\\
				GCL~\cite{chen2021joint} & 25.8 & 34.6 & 46.0 & 33.6 & 69.7 & 41.9\\
				\textbf{D$^2$CA (Ours)} & \textbf{35.5} & \textbf{42.7} & 50.6 & \textbf{34.0} & \textbf{70.3} & \textbf{46.6}\\
				\midrule
				\textit{Target}~\cite{yin2021self} & \textit{49.7} & \textit{69.8} & \textit{41.9} & \textit{46.5} & \textit{63.0} & \textit{54.2}\\
				\bottomrule
				\toprule
				\multicolumn{7}{c}{Source: \textbf{DISFA}. Target: \textbf{BP4D}. } \\
				\midrule
				\textit{Source}~\cite{yin2021self} & \textit{43.2} & \textit{38.3} & \textit{34.3} & \textit{54.6} & \textit{58.3} & \textit{45.7}\\
				\midrule
				DANN~\cite{ganin2015unsupervised} & 27.3 & 11.5 & 39.6 & 49.5 & 55.7 & 36.7\\
				MCD~\cite{saito2018maximum} & 31.1 & 24.0 & 32.6 & 66.5 & 54.5 & 41.7\\
				P-MCD~\cite{yin2021self} & 27.2 & 26.2 & 32.0 & 68.3 & 58.0 & 42.3\\
				\midrule
				TCAE~\cite{li2019self} & 39.2 & 44.7 & 32.8 & 68.5 & 57.5 & 48.5\\
				DGNet++~\cite{zou2020joint} & 33.6 & 35.7 & 41.0 & 68.6 & 65.2 & 48.8\\
				DRANet~\cite{lee2021dranet} & 45.1 & 49.7 & \textbf{52.6} & 60.9 & 51.1 & 51.9\\
				GCL~\cite{chen2021joint} & 49.9 & 48.8 & 44.1 & 62.7 & 46.5 & 50.4\\
				
				\textbf{D$^2$CA (Ours)} & \textbf{54.6} & \textbf{56.4} & 46.2 & \textbf{71.5} & \textbf{67.1} & \textbf{59.2}\\
				\midrule
				
				\textit{Target}~\cite{yin2021self} & \textit{55.1} & \textit{54.4} & \textit{46.6} & \textit{72.0} & \textit{86.3} & \textit{62.9}\\
				\bottomrule
			\end{tabular}
		}
	\end{center}
\end{table}

\begin{table}[htb]
	\caption{F1-scores of domain adaptation from BP4D+ to DISFA. \textbf{Bold} means the best.}
	\label{tab:bp4d_plus_disfa}
	\setlength\tabcolsep{5.0pt}
	\begin{center}
		\scalebox{0.9}{
			\begin{tabular}{c|c|c|c|c|c|c}
				\toprule
				AU & 1 & 2 & 4 & 6 & 12 & AVE\\
				\toprule
				\multicolumn{7}{c}{Source: \textbf{BP4D+}. Target: \textbf{DISFA}. } \\
				\midrule
				\textit{Source}~\cite{yin2021self} & \textit{2.4} & \textit{0.1} & \textit{14.1} & \textit{12.2} & \textit{27.2} & \textit{11.2} \\
				\midrule
				DANN~\cite{ganin2015unsupervised} & 33.4 & 22.7 & 43.2 & 24.0 & 35.8 & 31.8\\
				MCD~\cite{saito2018maximum} & 18.4 & 18.7 & 37.7 & 28.4 & 57.5 & 32.1\\
				P-MCD~\cite{yin2021self} & 25.2 & 10.9 & 56.5 & 37.0 & 59.5 & 37.8\\
				AFS~\cite{zhuo2023adaptive} & 45.2 & 35.1 & 28.6 &  39.7 & 49.4 & 39.6 \\
				CAF~\cite{xie2023collaborative} & 35.1 & 29.5 & 53.0 & 38.8 & 61.6 & 43.6 \\
				\midrule
				TCAE~\cite{li2019self} & 30.7 & 21.5 & 38.5 & 31.4 & 55.2 & 35.5\\
				DGNet++~\cite{zou2020joint} & 39.0 & 28.2 & 25.3 & 39.2 & 59.1 & 38.2\\
				DRANet~\cite{lee2021dranet} & 44.8 & 24.5 & 47.8 & \textbf{41.7} & 66.7 & 45.1\\
				GCL~\cite{chen2021joint} & 43.5 & 17.2 & 53.6 & 37.6 & 65.2 & 43.4\\
				\textbf{D$^2$CA (Ours)} & \textbf{45.4} & \textbf{36.5} & \textbf{59.4} & 39.9 & \textbf{67.2} & \textbf{49.7} \\
				\midrule
				\textit{Target}~\cite{yin2021self} & \textit{49.7} & \textit{69.8} & \textit{41.9} & \textit{46.5} & \textit{63.0} & \textit{54.2} \\
				\bottomrule
				\toprule
				\multicolumn{7}{c}{Source: \textbf{DISFA}. Target: \textbf{BP4D+}. } \\
				\midrule
				\textit{Source}~\cite{yin2021self} & \textit{22.3} & \textit{16.3} & \textit{15.8} & \textit{58.1} & \textit{65.4} & \textit{35.6}\\
				\midrule
				DANN~\cite{ganin2015unsupervised} & 14.7 & 9.8 & 11.1 & 61.1 & 67.9 & 32.9\\
				MCD~\cite{saito2018maximum} & 15.2 & 14.4 & 9.7 & 60.6 & 68.6 & 33.7\\
				P-MCD~\cite{yin2021self} & 15.6 & 14.7 & 17.7 & 60.7 & 74.5 & 36.6\\
				\midrule
				TCAE~\cite{li2019self} & 21.8 & \textbf{18.4} & 16.0 & 52.1 & 72.8 & 36.2\\
				DGNet++~\cite{zou2020joint} & 12.1 & 15.5 & 14.2 & 70.6 & 74.3 & 37.3\\
				DRANet~\cite{lee2021dranet} & 23.3 & 15.8 & 20.2 & 70.8 & 68.9 & 39.8\\
				GCL~\cite{chen2021joint} & 23.9 & 18.1 & 16.2 & 76.1 & 69.1 & 40.7\\
				
				\textbf{D$^2$CA (Ours)} & \textbf{24.3} & 17.8 & \textbf{20.6} & \textbf{77.6} & \textbf{76.8} & \textbf{43.4}\\
				\midrule
				
				\textit{Target}~\cite{yin2021self} & \textit{38.7} & \textit{30.7} & \textit{42.0} & \textit{85.2} & \textit{88.8} & \textit{57.1}\\
				\bottomrule
			\end{tabular}
		}
	\end{center}
\end{table}

\begin{table}[htb]
	\caption{F1-scores of domain adaptation from Aff-Wild2 to BP4D+. \textbf{Bold} means the best.}
	\label{tab:Aff-Wild2_bp4d_plus}
	\setlength\tabcolsep{2.0pt}
	\begin{center}
		\scalebox{0.9}{
			\begin{tabular}{c|c|c|c|c|c|c|c|c|c|c|c}
				\toprule
				AU & 1 & 2 & 4 & 6 & 7 & 10 & 12 & 15 & 23 & 24 & AVE\\
				\toprule
				\multicolumn{12}{c}{Source: \textbf{Aff-Wild2}. Target: \textbf{BP4D+}. } \\
				\midrule
				\textit{Source}~\cite{yin2021self} & \textit{13.9} & \textit{12.8} & \textit{13.4} & \textit{49.9} & \textit{55.1} & \textit{55.4} & \textit{56.8} & \textit{6.2} & \textit{10.1} & \textit{5.8} & \textit{27.9}\\
				
				\midrule
				DANN~\cite{ganin2015unsupervised} & 0.0 & 0.0 & 8.7 & 41.9 & 57.9 & 60.8 & 41.3 & 0.0 & 0.0 & 0.0 & 21.1\\
				MCD~\cite{saito2018maximum} & 1.4 & 3.2 & 8.9 & 44.0 & 52.2 & 53.0 & 48.4 & 8.0 & 12.7 & 2.9 & 23.5\\
				P-MCD~\cite{yin2021self} & 19.0 & 5.7 & 1.9 & 41.9 & 57.7 & 57.2 & 45.8 & 4.4 & 6.9 & 4.0 & 24.5\\
				\midrule
				TCAE ~\cite{li2019self} & 1.9 & 4.6 & 5.0 & 49.0 & 66.6 & 65.3 & 44.1 & 5.0 & 0.1 & 0.4 & 24.2\\
				DGNet++~\cite{zou2020joint} & 20.7 & 5.8 & 8.9 & 66.3 & 11.1 & 50.8 & 10.1 & 15.3 & 15.9 & \textbf{7.3} & 21.2\\
				DRANet~\cite{lee2021dranet} & 17.1 & 5.4 & 12.7 & 75.0 & 64.8 & 76.8 & 72.6 & 4.6 & 5.0 & 2.4 & 33.6\\
				GCL~\cite{chen2021joint} & \textbf{25.3} & 9.2 & 14.4 & 62.1 & 52.6 & 74.9 & 74.3 & \textbf{17.1} & 15.4 & 7.0 & 35.2\\
				
				\textbf{D$^2$CA (Ours)} & 19.2 & \textbf{15.7} & \textbf{14.7} & \textbf{76.6} & \textbf{76.1} & \textbf{80.0} & \textbf{76.2} & 15.3 & \textbf{20.8} & 6.7 & \textbf{40.1}\\
				\midrule
				
				\textit{Target}~\cite{yin2021self} & \textit{36.8} & \textit{28.1} & \textit{29.1} & \textit{84.9} & \textit{84.6} & \textit{88.5} & \textit{87.3} & \textit{39.4} & \textit{53.0} & \textit{26.2} & \textit{55.8}\\
				\bottomrule
				\toprule
				\multicolumn{12}{c}{Source: \textbf{BP4D+}. Target: \textbf{Aff-Wild2}. } \\
				\midrule
				\textit{Source}~\cite{yin2021self} & \textit{20.5} & \textit{17.3} & \textit{23.6} & \textit{27.1} & \textit{32.5} & \textit{27.2} & \textit{30.7} & \textit{1.9} & \textit{5.6} & \textit{2.3} & \textit{18.9}\\
				
				\midrule
				DANN~\cite{ganin2015unsupervised} & 11.6 & 5.9 & 0.2 & 29.9 & 47.9 & 39.7 & 22.4 & 1.2 & 5.2 & 0.0 & 16.4\\
				MCD~\cite{saito2018maximum} & 15.2 & 12.4 & 13.6 & 25.4 & 47.2 & 41.5 & 34.0 & 2.7 & 4.8 & 2.7 & 19.9\\
				P-MCD~\cite{yin2021self} & 10.4 & 6.9 & 7.7 & 32.4 & 53.6 & 43.1 & 37.4 & 6.9 & 5.0 & 2.7 & 20.6\\
				\midrule
				TCAE ~\cite{li2019self} & 22.4 & 12.5 & 25.1 & 38.9 & \textbf{53.9} & 43.7 & 26.8 & 6.2 & 4.9 & 2.7 & 23.7\\
				DGNet++~\cite{zou2020joint} & 21.2 & 15.5 & 21.2 & 29.1 & 33.1 & 31.5 & 30.8 & 5.1 & 4.9 & \textbf{5.8} & 19.8\\
				DRANet~\cite{lee2021dranet} & 25.3 & 15.5 & 18.8 & 26.8 & 48.4 & 31.9 & 19.6 & 0.7 & 3.6 & 0.0 & 19.1\\
				GCL~\cite{chen2021joint} & 24.5 & \textbf{19.6} & 26.9 & 38.6 & 45.0 & 14.9 & 38.1 & 3.4 & 3.9 & 4.5 & 21.9\\
				
				\textbf{D$^2$CA (Ours)} & \textbf{26.2} & 18.9 & \textbf{27.1} & \textbf{41.0} & 50.5 & \textbf{44.6} & \textbf{38.7} & \textbf{7.1} & \textbf{6.3} & 4.3 & \textbf{26.5}\\
				\midrule
				
				\textit{Target}~\cite{yin2021self} & \textit{49.3} & \textit{38.9} & \textit{39.8} & \textit{59.3} & \textit{72.1} & \textit{71.9} & \textit{68.4} & \textit{17.7} & \textit{12.9} & \textit{14.8} & \textit{44.5}\\
				\bottomrule
			\end{tabular}
		}
	\end{center}
\end{table}

\begin{table}[htb]
	\caption{F1-scores of domain adaptation from Aff-Wild2 to BP4D. \textbf{Bold} means the best.}
	\label{tab:Aff-Wild2_bp4d}
	\setlength\tabcolsep{2.0pt}
	\begin{center}
		\scalebox{0.9}{
			\begin{tabular}{c|c|c|c|c|c|c|c|c|c|c|c}
				\toprule
				AU & 1 & 2 & 4 & 6 & 7 & 10 & 12 & 15 & 23 & 24 & AVE\\
				\toprule
				\multicolumn{12}{c}{Source: \textbf{Aff-Wild2}. Target: \textbf{BP4D}. } \\
				\midrule
				\textit{Source}~\cite{yin2021self} & \textit{29.3} & \textit{19.9} & \textit{27.3} & \textit{62.8} & \textit{57.6} & \textit{59.1} & \textit{55.6} & \textit{17.5} & \textit{8.2} & \textit{5.4} & \textit{34.3}\\
				
				\midrule
				DANN~\cite{ganin2015unsupervised} & 0.0 & 0.0 & 0.1 & 42.5 & 58.4 & 61.2 & 54.1 & 0.3 & 0.0 & 0.0 & 21.7\\
				MCD~\cite{saito2018maximum} & 5.7 & 1.8 & 6.0 & 47.1 & 54.3 & 52.9 & 52.8 & 4.1 & 2.4 & 2.7 & 23.0\\
				P-MCD~\cite{yin2021self} & 25.3 & 16.2 & 15.7 & 41.4 & 57.6 & 59.4 & 50.1 & 6.4 & 7.6 & \textbf{10.3} & 29.0\\
				\midrule
				TCAE ~\cite{li2019self} & 0.4 & 0.2 & 6.7 & 52.7 & 65.0 & 71.4 & 49.0 & 0.0 & 0.0 & 0.0 & 24.5\\
				DGNet++~\cite{zou2020joint} & 21.6 & 15.5 & 28.8 & 31.1 & 36.5 & 35.9 & 25.3 & 25.5 & 21.9 & 8.5 & 25.1\\
				DRANet~\cite{lee2021dranet} & 29.6 & 17.3 & 29.7 & 64.1 & \textbf{66.1} & 72.5 & 75.7 & 16.4 & 9.6 & 0.0 & 38.1\\
				GCL~\cite{chen2021joint} & 37.4 & 24.1 & 30.1 & 70.5 & 63.8 & 69.8 & 67.0 & 15.0 & 20.5 & 8.2 & 40.6\\
				
				\textbf{D$^2$CA (Ours)} & \textbf{42.7} & \textbf{29.4} & \textbf{32.5} & \textbf{71.9} & 63.4 & \textbf{79.3} & \textbf{82.4} & \textbf{28.6} & \textbf{22.8} & 8.1 & \textbf{46.1}\\
				\midrule
				
				\textit{Target}~\cite{yin2021self} &  \textit{52.1} & \textit{46.7} & \textit{41.2} & \textit{76.4} & \textit{75.5} & \textit{82.6} & \textit{88.8} & \textit{39.8} & \textit{38.9} & \textit{46.4} & \textit{58.8}\\
				\bottomrule
				\toprule
				\multicolumn{12}{c}{Source: \textbf{BP4D}. Target: \textbf{Aff-Wild2}. } \\
				\midrule
				\textit{Source}~\cite{yin2021self} & \textit{16.1} & \textit{12.6} & \textit{19.2} & \textit{19.5} & \textit{46.8} & \textit{32.3} & \textit{29.2} & \textit{4.5} & \textit{2.4} & \textit{1.8} & \textit{18.4}\\
				
				\midrule
				DANN~\cite{ganin2015unsupervised} & 14.8 & 12.1 & 16.5 & 27.7 & 34.3 & 38.3 & 33.3 & 2.8 & 2.7 & 1.8 & 18.4\\
				MCD~\cite{saito2018maximum} & 15.0 & 10.4 & 13.6 & 23.8 & 34.4 & 34.3 & 29.5 & 1.9 & 2.2 & 2.2 & 16.7\\
				P-MCD~\cite{yin2021self} & 7.1 & 6.2 & 12.4 & 28.6 & 41.2 & 31.8 & 26.8 & 4.9 & 3.8 & 1.7 & 16.4\\
				\midrule
				TCAE ~\cite{li2019self} & 23.8 & 10.8 & 26.8 & 36.7 & 51.8 & \textbf{46.6} & 33.8 & 5.2 & 1.0 & 5.2 & 24.2\\
				DGNet++~\cite{zou2020joint} & 22.3 & 19.7 & 22.6 & 33.4 & 49.8 & 42.6 & 34.7 & 6.0 & 4.4 & \textbf{6.2} & 24.2\\
				DRANet~\cite{lee2021dranet} & 22.5 & \textbf{21.0} & 25.8 & 38.2 & 51.6 & 46.0 & 31.3 & 5.3 & 2.5 & 3.6 & 24.8\\
				GCL~\cite{chen2021joint} & 23.9 & 17.2 & 26.0 & 34.0 & 40.2 & 23.1 & 34.7 & 5.9 & 4.6 & 3.6 & 21.3\\
				
				\textbf{D$^2$CA (Ours)} & \textbf{24.4} & 17.1 & \textbf{28.6} & \textbf{38.8} & \textbf{53.7} & 42.5 & \textbf{35.7} & \textbf{6.2} & \textbf{5.4} & 4.3 & \textbf{25.7}\\
				\midrule
				
				\textit{Target}~\cite{yin2021self} & \textit{49.3} & \textit{38.9} & \textit{39.8} & \textit{59.3} & \textit{72.1} & \textit{71.9} & \textit{68.4} & \textit{17.7} & \textit{12.9} & \textit{14.8} & \textit{44.5}\\
				\bottomrule
			\end{tabular}
		}
	\end{center}
\end{table}

\begin{table}[htb]
	\caption{F1-scores of domain adaptation from Aff-Wild2 to GFT. \textbf{Bold} means the best.}
	\label{tab:Aff-Wild2_gft}
	\setlength\tabcolsep{2.0pt}
	\begin{center}
			\begin{tabular}{c|c|c|c|c|c|c|c|c|c|c}
				\toprule
				AU & 1 & 2 & 4 & 6 & 10 & 12 & 15 & 23 & 24 & AVE\\
				\toprule
				\multicolumn{11}{c}{Source: \textbf{Aff-Wild2}. Target: \textbf{GFT}. } \\
				\midrule
				\textit{Source}~\cite{yin2021self} & \textit{4.6} & \textit{14.1} & \textit{6.2} & \textit{46.8} & \textit{45.3} & \textit{50.3} & \textit{15.3} & \textit{21.6} & \textit{8.2} & \textit{23.6}\\
				
				\midrule
				DANN~\cite{ganin2015unsupervised} & 0.0 & 0.0 & 9.1 & 45.9 & 41.6 & 45.5 & 0.0 & 0.0 & 0.0 & 15.8\\
				MCD~\cite{saito2018maximum} & 5.9 & 6.1 & 8.6 & 35.0 & 37.5 & 38.3 & 2.5 & 22.0 & 8.3 & 18.3\\
				P-MCD~\cite{yin2021self} & 6.0 & 11.9 & 7.9 & 30.8 & 39.5 & 32.7 & 3.7 & 24.7 & 17.0 & 19.4\\
				\midrule
				TCAE ~\cite{li2019self} & 0.2 & 0.0 & \textbf{13.6} & 34.2 & 50.3 & 52.3 & 0.0 & 0.0 & 0.0 & 16.7\\
				DGNet++~\cite{zou2020joint} & 5.0 & \textbf{25.3} & 5.1 & 48.5 & 40.3 & 41.1 & 19.7 & 30.5 & 15.1 & 25.6\\
				DRANet~\cite{lee2021dranet} & 7.1 & 7.1 & 8.7 & 64.8 & 66.5 & 67.5 & 3.0 & 0.0 & 0.0 & 25.0\\
				GCL~\cite{chen2021joint} & 8.2 & 16.0 & 7.8 & 65.4 & 66.1 & 64.2 & 15.2 & 28.3 & 14.3 & 31.7\\
				
				\textbf{D$^2$CA (Ours)} & \textbf{10.0} & 24.9 & 9.3 & \textbf{67.2} & \textbf{67.6} & \textbf{77.9} & \textbf{22.6} & \textbf{35.5} & \textbf{17.1} & \textbf{36.9}\\
				\midrule
				\textit{Target}~\cite{yin2021self} & \textit{31.1} & \textit{39.6} & \textit{9.2} & \textit{77.7} & \textit{72.6} & \textit{78.7} & \textit{24.9} & \textit{41.5} & \textit{30.4} & \textit{45.1}\\
				\bottomrule
				\toprule
				\multicolumn{11}{c}{Source: \textbf{GFT}. Target: \textbf{Aff-Wild2}. } \\
				\midrule
				\textit{Source}~\cite{yin2021self} & \textit{13.0} & \textit{16.5} & \textit{17.5} & \textit{22.8} & \textit{22.8} & \textit{30.9} & \textit{3.7} & \textit{4.3} & \textit{3.4} & \textit{15.0}\\
				
				\midrule
				DANN~\cite{ganin2015unsupervised} & 8.3 & 15.1 & 6.8 & 25.2 & 26.8 & 27.2 & 2.9 & 2.7 & 4.7 & 13.3\\
				MCD~\cite{saito2018maximum} & 10.9 & 14.1 & 16.2 & 27.6 & 33.4 & 32.6 & 3.7 & 5.2 & 4.5 & 16.5\\
				P-MCD~\cite{yin2021self} & 8.6 & 14.2 & 17.5 & 28.7 & 40.3 & 31.5 & 5.4 & 4.9 & 4.8 & 17.3\\
				\midrule
				TCAE ~\cite{li2019self} & 6.4 & 13.0 & 7.4 & 28.2 & 40.1 & 39.0 & 6.6 & 4.0 & 3.4 & 16.5\\
				DGNet++~\cite{zou2020joint} & 17.4 & 17.8 & 16.5 & 16.3 & 28.1 & 21.8 & 5.6 & 4.8 & 4.3 & 14.7\\
				DRANet~\cite{lee2021dranet} & 18.6 & 19.5 & 18.5 & \textbf{31.7} & 29.5 & \textbf{39.1} & \textbf{7.1} & 5.4 & 2.0 & 19.0\\
				GCL~\cite{chen2021joint} & 24.7 & 18.8 & 22.2 & 16.4 & 38.3 & 32.7 & 3.7 & 4.9 & 4.6 & 18.5\\
				
				\textbf{D$^2$CA (Ours)} & \textbf{26.7} & \textbf{21.0} & \textbf{24.6} & 30.4 & \textbf{43.4} & 37.2 & 5.0 & \textbf{6.0} & \textbf{5.5} & \textbf{22.2}\\
				\midrule
				\textit{Target}~\cite{yin2021self} & \textit{46.6} & \textit{39.8} & \textit{27.2} & \textit{52.5} & \textit{68.3} & \textit{59.3} & \textit{14.4} & \textit{8.6} & \textit{8.1} & \textit{36.1}\\
				\bottomrule
			\end{tabular}
	\end{center}
\end{table}

\subsection{Doubly Contrastive Learning}
\label{sec:dcl}

With the above image or feature level constraints, they still cannot guarantee the complete feature decoupling and purification. In fact, information can freely leak between representations, e.g., all the AU and domain information can be merely encoded in the AU-relevant branch so that the decoder can easily synthesize the input facial images, leaving the domain features meaningless. To address this issue, we introduce a doubly contrastive learning strategy comprising image-level contrastive learning and feature-level contrastive learning. This strategy is specially designed to alleviate undesirable feature ambiguity and ensure the high quality of the altered facial images with transferred AUs.

\textbf{Image-level Contrastive Learning (ICL)}. We illustrate the  framework of ICL in Fig.~\ref{fig:dca_framework} (b).
During training, For each randomly sampled pair of source and target facial images, D$^2$CA will generate or reconstruct an additional set of 6 intermediate images. This set includes 2 images in the Universal Feature Purification stage, comprising the \textit{AU-altered source facial image} $\mathbf{I}_{s \rightarrow t}$ and \textit{AU-altered target facial image} $\mathbf{I}_{t \rightarrow s}$. It also involves 2 images in the Cyclical Feature Alignment with Source stage, namely the \textit{reconstructed source facial image} $\widehat{\mathbf{I}}_s$ and the \textit{reconstructed AU-altered source facial image} $\widehat{\mathbf{I}}_{s \rightarrow t}$.
Additionally, the generated image set consists of 2 images in the Cyclical Feature Alignment with Target stage, encompassing the \textit{reconstructed target facial image} $\widehat{\mathbf{I}}_t$ and the \textit{reconstructed AU-altered target facial image} $\widehat{\mathbf{I}}_{t \rightarrow s}$.

During the training process, we sample $N$ images from both the source and target domains, resulting in a total of $8N$ images involved in the image-level contrastive learning (ICL). In the case of a positive sample pair, such as ($\mathbf{I}_s, \widehat{\mathbf{I}}_s$), the remaining $8N-2$ samples are considered as negative samples. The optimization objective for ICL can be formulated as follows:
\begin{equation}
	\label{eq:icl}
	\mathcal{L}_{icl} =  -\text{log} \frac{\text{exp}(q \cdot k_{+} / \tau)}{\text{exp}(q \cdot k_{+}/ \tau) + \sum_{i=1}^{8N-2} \text{exp}( q \cdot k_{i} / \tau)},
\end{equation}
where $q$ and $k_{+}$ mean the features of the positive sample pair. $\tau$ is a temperature hyper-parameter used to scale the distribution of the compared features. 
To derive the features for image-level contrastive learning, we concatenate the AU and domain features of each image and employ two fully-connected layers for feature fusion and projection. By utilizing ICL, our proposed D$^2$CA is adept at discerning the subtle differences between the AU-altered facial images and the original input images, as well as the reconstructed ones, thereby preventing the mode collapse of the generated AU-altered facial images.

\textbf{Feature-level Contrastive Learning (FCL)}. In an effort to strengthen the feature decoupling and adaptation, we postulate that within the source dataset, the domain representations originating from the same identity but different facial images should exhibit greater similarity compared to those from different identities. This conjecture forms the basis for the development of a margin-based contrastive learning loss, aimed at reinforcing the feature purification and alignment throughout the FCL procedure.

As depicted in Fig.~\ref{fig:feature_cl}, the application of the margin-based CL loss is favored due to its ability to accommodate certain degrees of intra-subject variance. For instance, a series of faces derived from the same facial video might display varying degrees of pose variations. With this consideration, a margin loss is defined as follows:
\begin{equation}
	\label{eq:fcl}
	\mathcal{L}_{fcl} = \frac{1}{|S|} \sum_{(i,j,k) \in S} \text{max}(0, \alpha - \text{cos}(\phi_i^s, \phi_j^s) + \text{cos}(\phi_i^s, \phi_k^s))
\end{equation}
where we collect a triple tuple set $S=\{(i,j,k)| id[i] = id[j], id[i] \neq id[k] \}$. The $id[i]$ means the identity of sample $i$, $cos(\cdot,\cdot)$ means the cosine similarity between two feature vectors. The loss in Equ.~\ref{eq:fcl} enforces the domain features, whether belonging to the different identities or not, to maintain a difference within the margin $\alpha$. This constraint serves the purpose of preventing the emergence of trivial domain features. Simultaneously, FCL indirectly encourages the shared AU encoder to acquire domain-invariant features. 

In summary, we have introduced the incorporation of feature and image level consistency constraints in the Cyclical Feature Alignment with Source/Target stages, aiming to achieve feature decoupling. Additionally, to attain feature alignment and adaptation, we have included supervised AU loss on the AU features and a margin-based contrastive loss on the domain features. Moreover, for the concatenation of the AU and domain features, we have utilized the instance-level contrastive loss to further ensure feature decoupling and purification. The constraints embedded within our proposed method have been designed to ensure feature purification and adaptation effects at both the global and local scales.

\subsection{Objective Optimization}
We integrate basic feature decoupling loss in Equ.~\ref{equ:decoupling} and the doubly contrastive loss in Equ.~\ref{eq:icl}$\sim$Equ.~\ref{eq:fcl} to reach the full objective:
\begin{align}
	\mathcal{L}_{\text{tot}} = \mathcal{L}_{deco} + \lambda (\mathcal{L}_{icl} + \mathcal{L}_{fcl}) ,
\end{align}
where $\mathcal{L}_{deco}$ is the basic feature decoupling loss defined in Equ.~\ref{equ:decoupling}. $\lambda$ controls the importance of DCL.

\begin{table*}
	\caption{Ablation study of D$^2$CA with and without image-/feature-level contrastive learning mechanism. Average F1-scores of domain adaptation are reported.}
	\label{tab:ablation_study}
	\begin{center}
		\scalebox{0.95}{
			\begin{tabular}{c|c|c|c|c|c|c|c|c}
				\toprule
				Model & B+ $\rightarrow$ D & D $\rightarrow$ B+ & B+ $\rightarrow$ G & G $\rightarrow$ B+ & A $\rightarrow$ B+ & B+ $\rightarrow$ A & A $\rightarrow$ G & G $\rightarrow$ A \\
				\midrule
				D$^2$CA (\textit{w/o ICL, FCL}) & 37.3 & 32.3 & 24.1 & 29.2 & 32.4 & 20.4 & 23.5 & 15.9\\
				D$^2$CA (\textit{w/o FCL}) & 46.7 & 40.5 & 34.3 & 36.8 & 37.7 & 24.5 & 33.2 & 19.1\\
				\textbf{D$^2$CA} & \textbf{49.7} & \textbf{43.4} & \textbf{37.8} & \textbf{39.5} & \textbf{40.1} & \textbf{26.5} & \textbf{36.9} & \textbf{22.2}\\
				\bottomrule
			\end{tabular}
		}
	\end{center}
\end{table*}

\begin{table*}
	\caption{Frechet inception distance (FID) for various D$^2$CA designs (lower is better).}
	
	\label{tab:fid_score}
	\begin{center}
		\scalebox{0.95}{
			\begin{tabular}{c|c|c|c|c|c|c|c|c}
				\toprule
				Model & B+ $\rightarrow$ D & D $\rightarrow$ B+ & B+ $\rightarrow$ G & G $\rightarrow$ B+ & A $\rightarrow$ B+ & B+ $\rightarrow$ A & A $\rightarrow$ G & G $\rightarrow$ A \\
				\midrule
				D$^2$CA (\textit{w/o ICL, FCL}) & 66.6 & 73.9 & 32.9 & 71.3 & 65.7 & 41.9 & 44.7 & 43.3\\
				D$^2$CA (\textit{w/o FCL}) & 57.7 & 48.7 & 30.1 & 48.4 & 50.5 & 41.7 & 38.6 & 42.4\\
				\textbf{D$^2$CA} & \textbf{46.9} & \textbf{37.7} & \textbf{29.5} & \textbf{44.4} & \textbf{47.6} & \textbf{30.4} & \textbf{29.2} & \textbf{30.1}\\
				\bottomrule
			\end{tabular}
		}
	\end{center}
\end{table*}

\section{Experiments}
\textbf{Datasets.} We evaluate the proposed D$^2$CA on five publicly available  benchmarks for AU detection, including BP4D+~\cite{zhang2016multimodal}, BP4D~\cite{zhang2014bp4d}, DISFA~\cite{mavadati2013disfa}, GFT~\cite{girard2017sayette}, Aff-Wild2~\cite{kollias2019expression}.
\textbf{BP4D} contains 41 participants (23 females and 18 males). There are
about 146000 frames with available 12 AU labels. For within-domain evaluation, we split the dataset into 3 folds based on subject IDs and conduct a 3-fold cross-validation.
\textbf{BP4D+}  is a multi-modal spontaneous emotion dataset consisting of 140 subjects. There are around 198,000 frames annotated with the same 12 AUs as that in BP4D.
\textbf{DISFA} consists of 27 participants. Approximately 130,000 frames were annotated with 8 AUs. The AUs are labelled with intensities from 0 to 5. The frames with intensities greater than 1 were considered as positive, while others were treated as negative. For within-domain evaluation, we use the subject-independent 3-fold cross-validation protocol.
\textbf{GFT} consists of 96 participants from 32 three-person groups with natural social interaction. Each subject has about 1,800 frames annotated with 10 AUs.
The moderate out-of-plane head motion make AU detection challenging for GFT dataset. 
\textbf{Aff-Wild2} is annotated on a per-frame basis, encompassing the seven fundamental expressions, namely happiness, surprise, anger, disgust, fear, sadness, and the neutral state, along with 12 action units (AU1, AU2, AU4, AU6, AU7, AU10, AU12, AU15, AU23, AU24, AU25, AU26) and the metrics of valence and arousal. The dataset comprises a total of 564 videos, comprising approximately 2.8 million frames, with the participation of 554 subjects, consisting of 326 male and 228 female individuals. Specifically, the publicly available training dataset contains 1359690 frames involving 295 subjects. The validation set consists of 445845 frames involving 105 subjects. Here we use the validation set for evaluation.

\textbf{Training and evaluation protocol.} 
In our cross-domain experiments, we adhere to the methodology outlined in~\cite{yin2021self}, wherein we designate BP4D+ and BP4D as the source/target domains and DISFA or GFT as the target/source domain. Furthermore, we utilize a large-scale ``in-the-wild'' database Aff-wild2 as both the source and the target domain interchangeably for further evaluations.
During training, we exploit the labeled source images and unlabeled target
images for unsupervised cross-domain adaptation. 
We use the official partitions provided with GFT and partition DISFA into three subject-independent folds for training/validation/testing.
For evaluation on BP4D and DISFA, we use the five shared AUs: AU1, AU2, AU4, AU6, AU12.
For evaluation on BP4D and GFT, we use the ten shared AUs: AU1, AU2, AU4, AU6, AU10, AU12, AU14, AU15, AU23, AU24.  
We use the F1 score ($F1=\frac{2RP}{R+P}$) to measure the performance of the compared methods, where $R$ and $P$ denote recall and precision, respectively. In addition, we compute the average performance over all AUs (AVE). 

\begin{figure}[hb]
	\centering
	\includegraphics[width=1.0\linewidth]{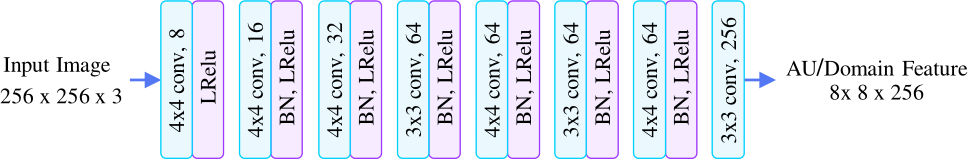}
	\caption{
		Network configuration of the shared AU encoder $\mathcal{E}_{au}$. Typically, the private domain encoders utilize the same neural network structure.
	}
	\label{fig:net_encoder}
\end{figure}

\begin{figure}[htb]
	\centering
	\includegraphics[width=1.0\linewidth]{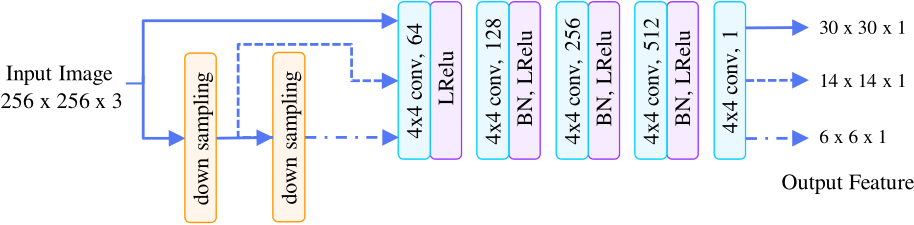}
	\caption{
		Network configuration of the image discriminator $\mathcal{D}^s$ and $\mathcal{D}^t$.	
	}
	\label{fig:net_discriminator}
\end{figure}

\begin{figure*}[htb]
	\centering
	\includegraphics[width=0.9\linewidth]{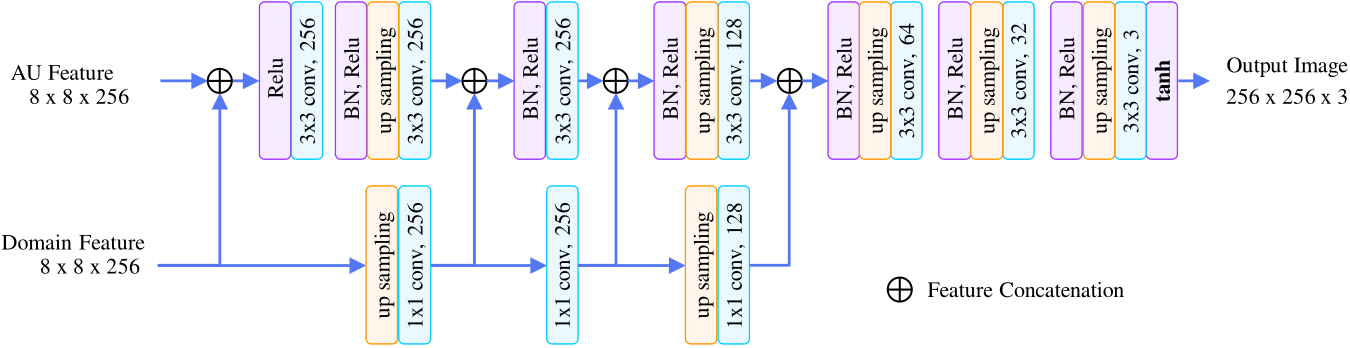}
	\caption{
		Network configuration of the image decoder $\mathcal{G}$.
	}
	\label{fig:net_decoder}
\end{figure*}

\textbf{Implementation details.} The optimal setting for $\lambda$, $\gamma_1$, $\gamma_2$, $\gamma_3$, $\gamma_4$ was set as 0.1, 1.0, 5.0, 0.1, 1.0 according to the cross-domain AU detection performance on the validation set. We set hyper-parameter $\tau=0.07$ and $\alpha=0.1$. We implemented all the experiments using PyTorch on two RTX 3090 GPUs, each with 24 GB
memory. We set the respective training batch size $N=128$ for the source and target images. 
We set the learning rate as $0.001$ and trained D$^2$CA for 50 epochs until convergence.
Additionally, we present the detailed network architectures in D$^2$CA, including the shared AU encoder $\mathcal{E}_{au}$ in Fig.~\ref{fig:net_encoder}; the private source/target domain encoder $\mathcal{E}_{dm}^s$/$\mathcal{E}_{dm}^t$ in Fig.~\ref{fig:net_encoder}; the image discriminator $\mathcal{D}^s$/$\mathcal{D}^t$ in Fig.~\ref{fig:net_discriminator}; image decoder $\mathcal{G}$ in Fig.~\ref{fig:net_decoder}.

As can be seen, the shared AU encoder $\mathcal{E}_{au}$ has the same network structure with the private domain encoders. The encoder is composed of eight convolutional layers, containing 0.35 million parameters, and requires 82.59 million floating point operations (FLOPs) per inference. The inference time is 2.8 milliseconds.
The image discriminator $\mathcal{D}^s$/$\mathcal{D}^t$ in Fig.~\ref{fig:net_discriminator} is patchGAN-based~\cite{zou2020joint}. The image discriminators take the original and synthesized face images as input and extract multi-scale features for fine-grained feature discrimination. The image decoder $\mathcal{G}$ in Fig.~\ref{fig:net_decoder} consists of seven convolutional layers and an auxiliary branch that repeatedly incorporates the domain features via feature concatenation.

\begin{figure*}[htb]
	\centering
	\includegraphics[width=0.9\linewidth]{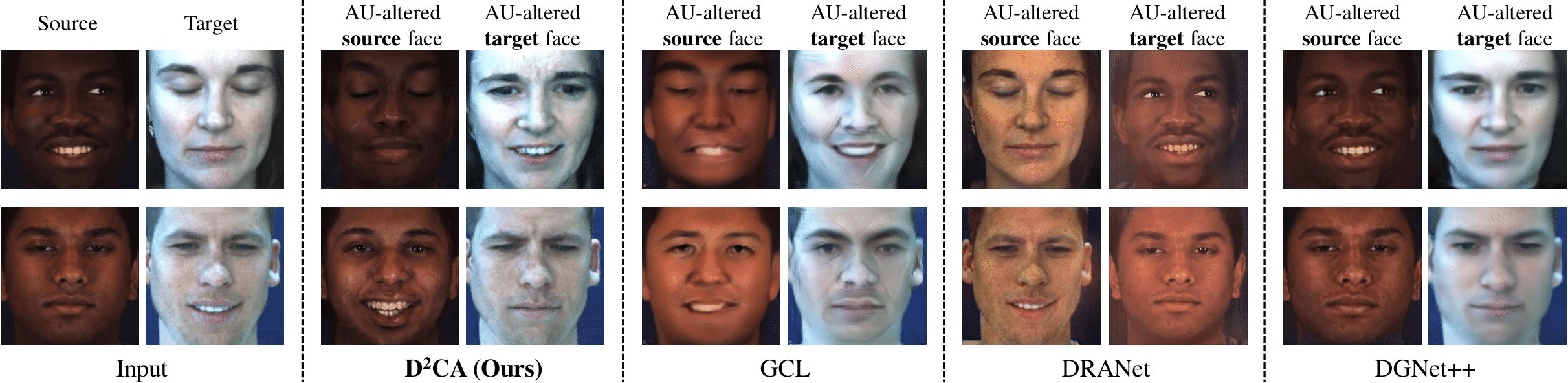}
	\caption{
		Comparison of the generated face images across two cross-domains (Source: BP4D+. Target: DISFA).
	}
	\label{fig:compared_method}
\end{figure*}

\begin{figure*}[htb]
	\centering
	\includegraphics[width=1.0\linewidth]{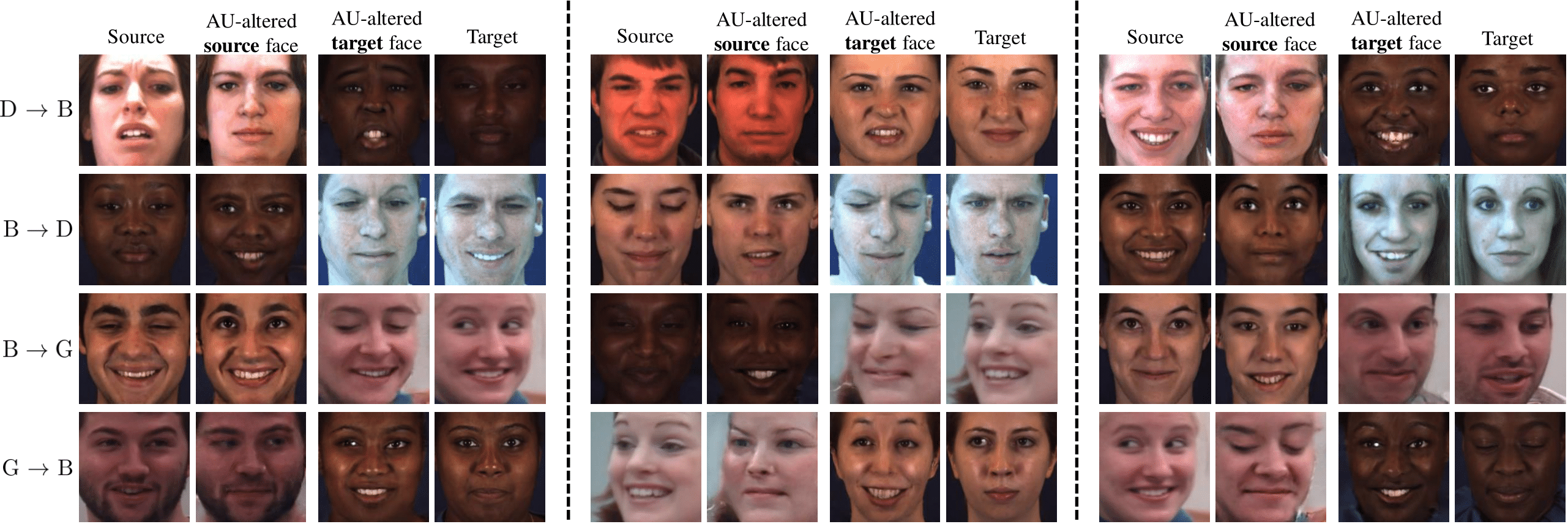}
	\caption{
		Examples of cross-domain synthesized face images by our proposed D$^2$CA.
		From 1st to 4th rows, D$^2$CA consistently generates visually pleasing AU-changed faces, indicating the AU-relevant/-irrelevant features are well separated. 
		B, D, and G represent BP4D, DISFA and GFT, respectively.
	}
	\label{fig:face_vis_bp4d}
\end{figure*}

\begin{figure*}[htb]
	\centering
	\includegraphics[width=1.0\linewidth]{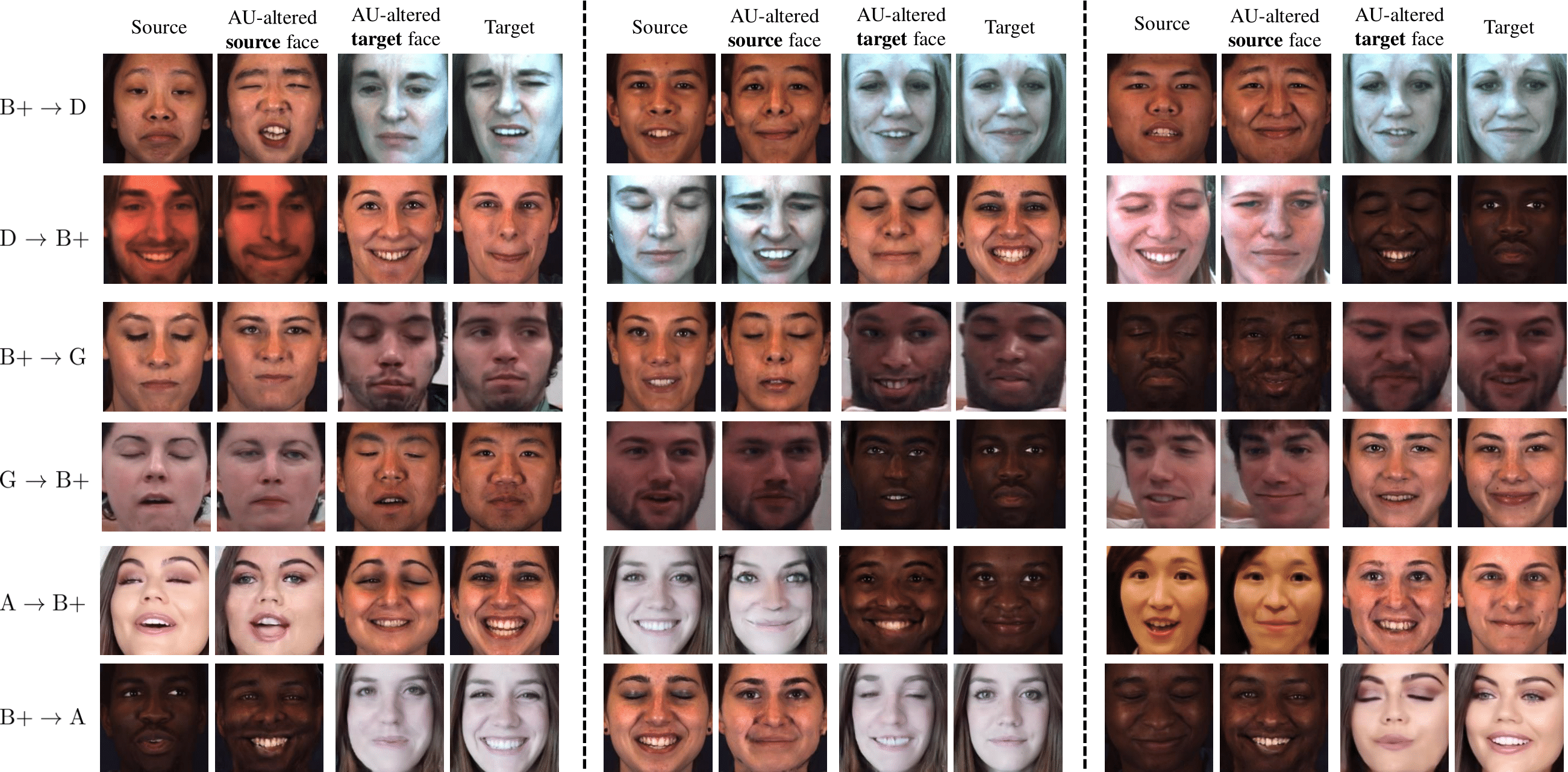}
	\caption{
		Examples of cross-domain synthesized face images by our proposed D$^2$CA.
		B+, B, D, G and A represent BP4D+, BP4D, DISFA, GFT and Aff-wild2 respectively.
	}
	\label{fig:face_vis_bp4d_plus}
\end{figure*}

\begin{figure*}[htb]
	\centering
	\includegraphics[width=1.0\linewidth]{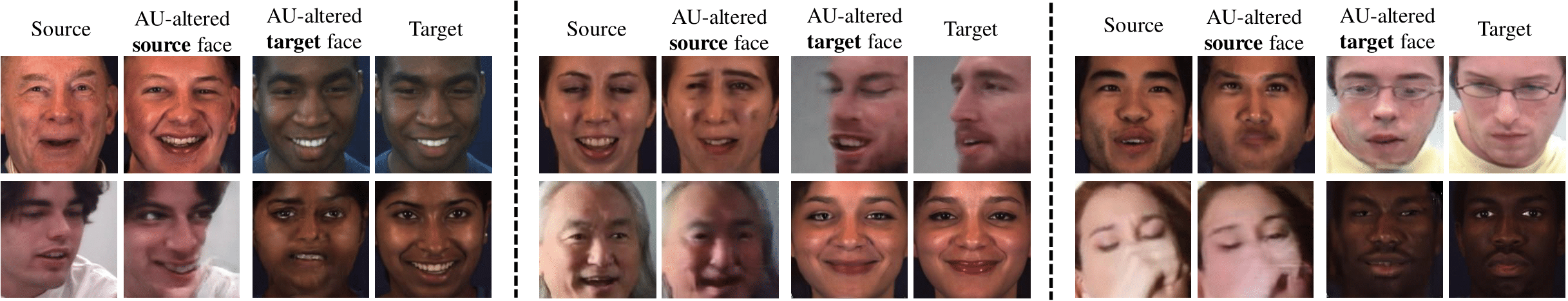}
	\caption{
		Examples of cross-domain synthesized failed cases face images by D$^2$CA. They suggest D$^2$CA is susceptible to heavy domain discrepancies.
	}
	\label{fig:face_vis_failed_case}
\end{figure*}

\subsection{Comparison with the State-of-the-arts}
We compare D$^2$CA with the representative UDA methods for AU detection,  including DANN~\cite{ganin2015unsupervised}, MCD~\cite{saito2018maximum}, P-MCD~\cite{yin2021self}, Adaptive Feature Swapping (AFS)~\cite{zhuo2023adaptive}, Collaborative Alignment Framework (CAF)~\cite{xie2023collaborative}.
We also compare D$^2$CA with three representative feature-decoupling-based UDA methods, including DG-Net++~\cite{zou2020joint}, DRANet~\cite{lee2021dranet}, GCL~\cite{chen2021joint}. Notably, DG-Net++ and GCL were originally developed for cross-domain person re-identification but were adapted for cross-domain AU detection based on their feature disentangling principle. They can be naturally used for cross-domain AU detection following their feature disentangling principle. For a fair comparison with D$^2$CA,  we re-trained these three models using the FACS datasets with the released codes.
We also compare D$^2$CA with the self-supervised AU detection method TCAE~\cite{li2019self}, where a pre-trained AU detection model is publicly available. Following TCAE~\cite{li2019self}, we train a linear classifier with the pre-trained model and evaluate its generalization capability on the target FACS dataset.

We show the quantitative cross-domain AU detection comparison in Tab.~\ref{tab:bp4d_gft}, Tab.~\ref{tab:bp4d_plus_gft}, Tab.~\ref{tab:bp4d_disfa}, Tab.~\ref{tab:bp4d_plus_disfa}, Tab.~\ref{tab:Aff-Wild2_bp4d_plus}, Tab.~\ref{tab:Aff-Wild2_bp4d}, Tab.~\ref{tab:Aff-Wild2_gft}.
In all the tables, we use the ``Source'' and ``Target'' reported in ~\cite{yin2021self} as the indicators for the lower and upper bounds. ``Source'' means directly evaluating the
source model on the target domain. ``Target'' means within-domain training and evaluation with the target data. 
Obviously, our proposed D$^2$CA obtains superior cross-domain AU detection performance than other compared methods.

Compared with the general DA methods including DANN~\cite{ganin2015unsupervised}, MCD~\cite{saito2018maximum} and P-MCD~\cite{yin2021self}, AFS~\cite{zhuo2023adaptive}, CAF~\cite{xie2023collaborative}, our proposed D$^2$CA obtains consistent improvements in the average F1 scores.  The observed improvements can be attributed to the fact that the compared general DA methods function within the global feature space, wherein AU-relevant and AU-irrelevant factors are intertwined. Consequently, the adaptation of AU-relevant features is unavoidably encumbered by the presence of AU-unrelated features.

A conspicuous observation is the discernible pattern that, in employing the GFT as the target dataset, the source dataset of BP4D+ manifests more substantial enhancements compared to those observed in the context of BP4D. Analogously, a similar trend emerges when DISFA serves as the target dataset. Given BP4D+'s inclusion of a larger cohort of subjects and training frames, the observed improvements underscore the advantageous impact of incorporating a greater quantity of source training images for the facilitation of UDA.

Compared with the feature-decoupling-based UDA methods, D$^2$CA consistently shows its benefits in Tab.\ref{tab:bp4d_gft}$\sim$Tab.\ref{tab:Aff-Wild2_gft}.
We conjecture they cannot fully separate the AU-relevant/-irrelevant factors as AU represents fine-grained facial deformation and the AU-relevant representation can not be separated using the classical style-content separation paradigm. 
For a further investigation, we synthesize several AU-changed faces using the randomly sampled source and target faces and illustrate the comparison in Fig.~\ref{fig:compared_method}. D$^2$CA consistently generates visually pleasing AU-altered source and target facial images. As a comparison, the synthesized facial images in DGNet++, DRANet and GCL do not effectively reflect the AU-altered but domain-unchanged property, suggesting that the features are not sufficiently disentangled.
D$^2$CA also outperforms TCAE whose encoded AU features are irrelevant from facial pose. This is reasonable because the domain component compromises various factors including identity, race, illumination, et al. Compared with TCAE, D$^2$CA fosters a less intertwined representation, thus allowing the domain component to encompass different facets of variation.

\begin{figure*}[htb]
	\centering
	\includegraphics[width=1.0\linewidth]{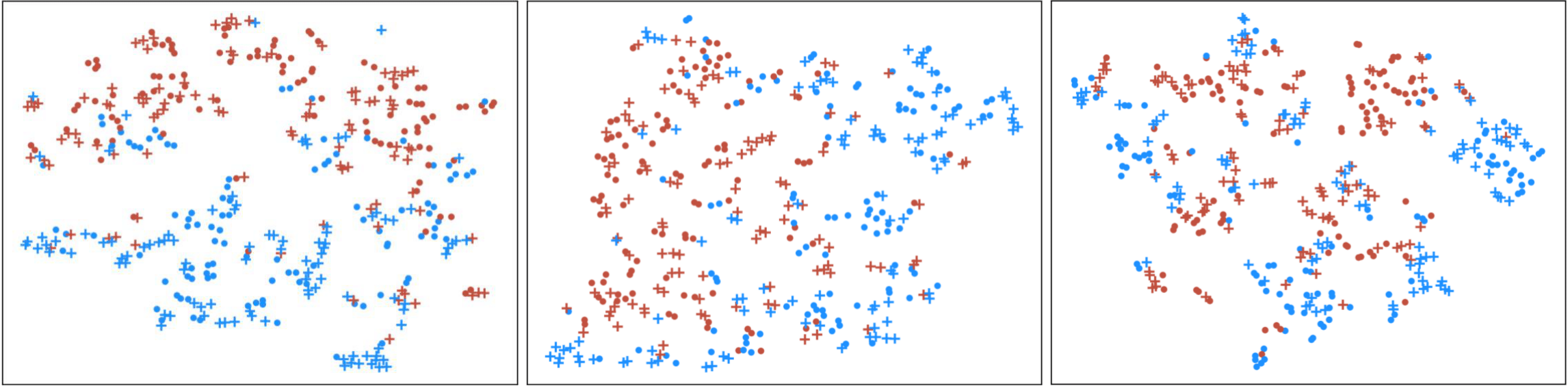}
	\caption{
		t-SNE visualization of the decoupled AU-irrelevant features. D$^2$CA shows the best feature discrimination w.r.t AU6. $\bullet$ means BP4D samples. $+$ means the DISFA samples. \textcolor{redcolor}{Red}/\textcolor{bluecolor}{Blue} indicate AU6 exists or not. Left: D$^2$CA. Middle: D$^2$CA (\textit{w/o FCL}). Right: D$^2$CA (\textit{w/o ICL, FCL}).
	}
	\label{fig:featrue_visualize}
\end{figure*}

\begin{figure}[htb]
	\centering
	\includegraphics[width=1.0\linewidth]{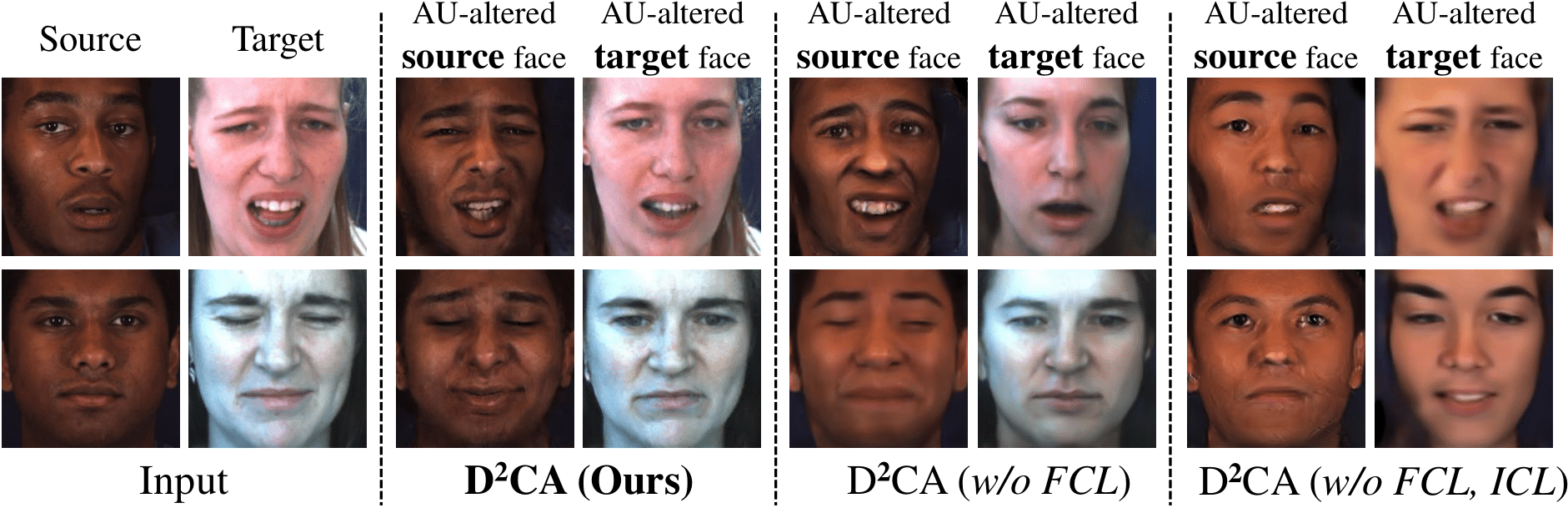}
	\caption{
		Ablation study of the cross-domain synthesized facial images (Source: BP4D+. Target: DISFA).
	}
	\label{fig:abla_face}
\end{figure}

\subsection{Analysis and Ablation study}
\textbf{Cross-domain synthesized facial images.}  We show various cross-domain generated facial images in Fig.~\ref{fig:face_vis_bp4d}, Fig.~\ref{fig:face_vis_bp4d_plus} and Fig.~\ref{fig:face_vis_failed_case}. Each row denotes a different source or target FACS dataset combination.
It is important to note that cross-domain facial image synthesis presents significant challenges owing to prominent domain gaps and limited data diversity, particularly with the FACS datasets, which comprise only a limited number of identities. 
Fig.~\ref{fig:face_vis_bp4d} and Fig.~\ref{fig:face_vis_bp4d_plus} verify that D$^2$CA is able to generate visually pleasing AU-altered facial images over different domain pairs. As illustrated, the source and target image pairs may illustrate diverse skin tones, poses and illuminations. The results indicate the AU-relevant/-irrelevant features in our proposed D$^2$CA are well separated.

We additionally illustrate several representative failed cases in Fig.~\ref{fig:face_vis_failed_case}. As can be seen, D$^2$CA cannot separate the AU-relevant/-irrelevant features when the domain shift is heavy. For instance, the limited representation of aged identities within the BP4D+ dataset exacerbates the challenges, as depicted in the first column and first row. Furthermore, the complexities associated with extreme head poses are illustrated in the second column of the first row, while the third column underscores the intricate nature of dealing with facial occlusions entangled with pose variations. Mitigating the intricacies of cross-domain AU detection within the realm of substantial domain discrepancies remains a persistent challenge, necessitating continued exploration in future research pursuits.

\textbf{Quantitative analysis.} We evaluate the effects of
D$^2$CA’s key components, including feature-level CL (FCL) and image-level CL (ICL).
Firstly, we show the Fréchet inception distance (FID)~\cite{heusel2017gans}  of D$^2$CA (\textit{w/o ICL, FCL}),  D$^2$CA (\textit{w/o FCL}) and D$^2$CA  in Tab.~\ref{tab:fid_score}.
D$^2$CA (\textit{w/o ICL, FCL}) denotes D$^2$CA without the doubly contrastive learning paradigm. D$^2$CA (\textit{w/o FCL}) means D$^2$CA without feature-level contrastive learning paradigm. 
FID is commonly used to measure the distribution similarity of generated images to the real images. In Tab.~\ref{tab:fid_score}, 10,000 generated images and 10,000 training images were randomly sampled from the initial dataset for the purpose of FID computation.
We observe that D$^2$CA obtains consistent FID improvements than D$^2$CA (\textit{w/o ICL, FCL}) or D$^2$CA (\textit{w/o FCL}). This is in accordance with the observations in Fig.~\ref{fig:abla_face}, where D$^2$CA-generated cross-domain facial images are more vivid than its counterparts.
Secondly, we show the cross-domain AU detection accuracy of D$^2$CA (\textit{w/o ICL, FCL}),  D$^2$CA (\textit{w/o FCL}) and D$^2$CA in Tab.~\ref{tab:ablation_study}.
The comparisons in Tab.~\ref{tab:ablation_study} verify that without the doubly contrastive learning paradigm, the cross-domain AU detection shows obvious degradation. Furthermore, the comparison between D$^2$CA (\textit{w/o ICL, FCL}) and D$^2$CA (\textit{w/o FCL}) indicates ICL helps the model lift the performance consistently. Finally, D$^2$CA with the doubly contrastive learning mechanism obtains the best results. As the generated AU-changed faces imply how well the AU-relevant/-irrelevant features are decoupled, we illustrate them next.

\textbf{Qualitative analysis.} 
As shown in Fig.~\ref{fig:abla_face}, D$^2$CA shows the best visual quality for the generated AU-altered source or target facial images.
The comparison between D$^2$CA (\textit{w/o FCL, ICL}) and D$^2$CA (\textit{w/o FCL}) suggests ICL can help lift the quality of the generated face images.
Besides, the comparison between D$^2$CA (\textit{w/o FCL}) and D$^2$CA verifies FCL is able to strength the feature decoupling.
To further investigate the contributions of ICL and FCL, we visualize the encoded AU features of D$^2$CA, D$^2$CA (\textit{w/o ICL}), D$^2$CA (\textit{w/o ICL, FCL}) in Fig.~\ref{fig:featrue_visualize}.
Two hundred facial images were randomly sampled from the testing set of the target DISFA and source BP4D datasets. 
The features of the selected facial images are projected into a 2D space by t-SNE.
Obviously, D$^2$CA shows the best feature discrimination w.r.t AU6. The comparison in Fig.~\ref{fig:featrue_visualize} further verifies the effectiveness of the doubly contrastive learning paradigm.

\section{Conclusion}
Within this paper we e have introduced the Decoupled Doubly Contrastive Adaptation (D$^2$CA) methodology for cross-domain action unit detection, aiming to ensure AU features remain robust to domain variations. Our approach involves encoding and decoupling facial images into AU-relevant and irrelevant spaces, supplemented by cross-domain face synthesis and a cyclical feature alignment mechanism to enhance feature separation. To further improve feature alignment and adaptation, a doubly contrastive learning strategy is employed. Experimental results consistently demonstrate the effectiveness of D$^2$CA, but challenges persist under extreme domain discrepancies, and performance may degrade with limited data diversity. Future work will involve exploring the integration of vision foundation models to enhance robustness against severe domain shifts and extend the proposed approach to a broader range of facial analysis tasks.

\section{Acknowledgment}
This research was partially supported by the RIE2020 AME Programmatic Fund, Singapore (No. A20G8b0102), the A*STAR Prenatal / Early Childhood Grant (No. H22P0M0002), Strategic Priority Research Program of the Chinese Academy of Sciences under Grant XDB0680202, Natural Science Foundation of China (No. U2003207).

\bibliographystyle{IEEEtran} 
\bibliography{reference}

\begin{IEEEbiography}[{\includegraphics[width=1in,height=1.25in,clip,keepaspectratio]{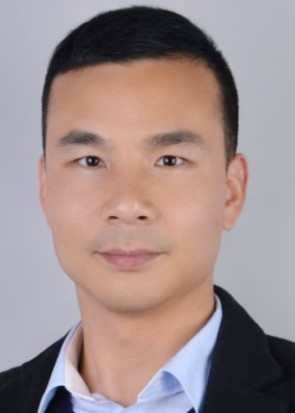}}]
{Yong Li} is currently an Associate Professor with the School of Computer Science and Engineering, Southeast University, China. He received the Ph.D. degree from Institute of Computing Technology (ICT), Chinese Academy of Sciences in 2020. He has previously held a position at Nanjing University of Science and Technology. In 2023-2024, he worked as a Research Fellow in City Universitiy of Hong Kong and Nanyang Technological University.  His research interests include face-related deep learning, human-centered affective computing. His research results have been expounded in more than 40 publications at prestigious journals and prominent conferences, such as IEEE TPAMI, IEEE TIP, IEEE TAC, IEEE TMM, NeurIPS, CVPR, ICCV.  For more information, please visit his personal website: https://mysee1989.github.io/.
\end{IEEEbiography}

\begin{IEEEbiography}[{\includegraphics[width=1in,height=1.25in,clip,keepaspectratio]{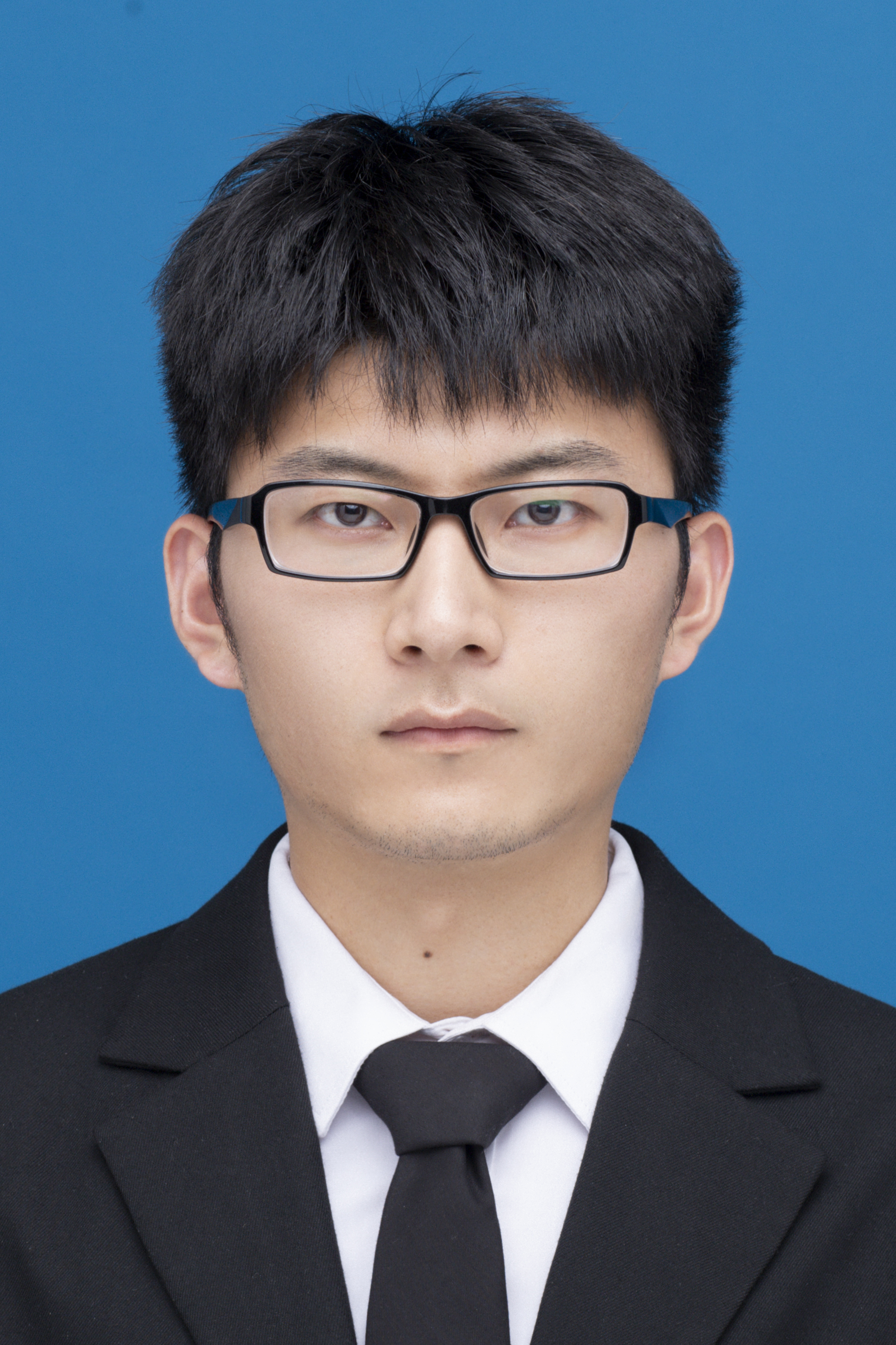}}]
	{Menglin Liu} received the B.E. degree from Nanjing University of Science and Technology, Nanjing, China in 2021. He is working towards the M.S. degree in computer science and technology. His research interests include computer vision, facial action unit detection and domain adaptation.
\end{IEEEbiography}

\begin{IEEEbiography}[{\includegraphics[width=1in,height=1.25in,clip,keepaspectratio]{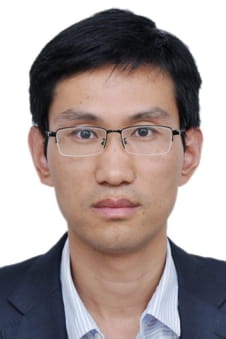}}]{Zhen Cui} received the Ph.D. degree from Institute of Computing Technology (ICT), Chinese Academy of Sciences in 2014. He was a Research Fellow in the Department of Electrical and Computer Engineering at National University of Singapore (NUS) from Sep 2014 to Nov 2015. He also spent half a year as a Research Assistant on Nanyang Technological University (NTU) from Jun 2012 to Dec 2012. Currently, he is a Professor of Nanjing University of Science and Technology, China. His research interests cover computer vision, pattern recognition and machine learning, especially focusing on  vision perception and computation, graph deep learning, etc.
\end{IEEEbiography}

\begin{IEEEbiography}
[{\includegraphics[width=1in,height=1.25in,clip,keepaspectratio]{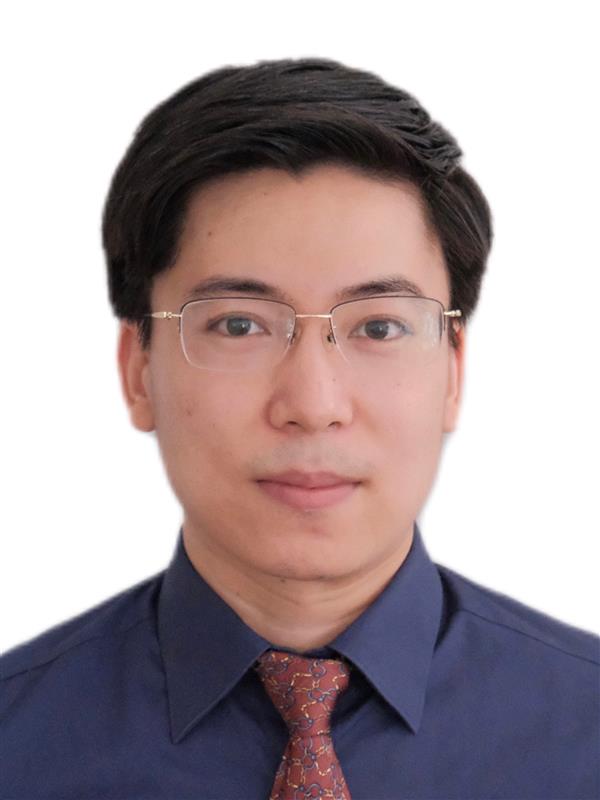}}]{Yi Ding}(Graduate student member, IEEE) earned a Ph.D. in Computer Science and Engineering from Nanyang Technological University, Singapore, in 2023, a master's degree in Electrical and Electronics Engineering from the same university in 2018, and a bachelor's degree in Information Science and Technology from Donghua University, Shanghai, China, in 2017. His research interests encompass brain-computer interface, deep learning, graph neural networks, neural signal processing, affective computing, and multimodal learning.
\end{IEEEbiography}

\begin{IEEEbiography}[{\includegraphics[width=1in,height=1.25in,clip,keepaspectratio]{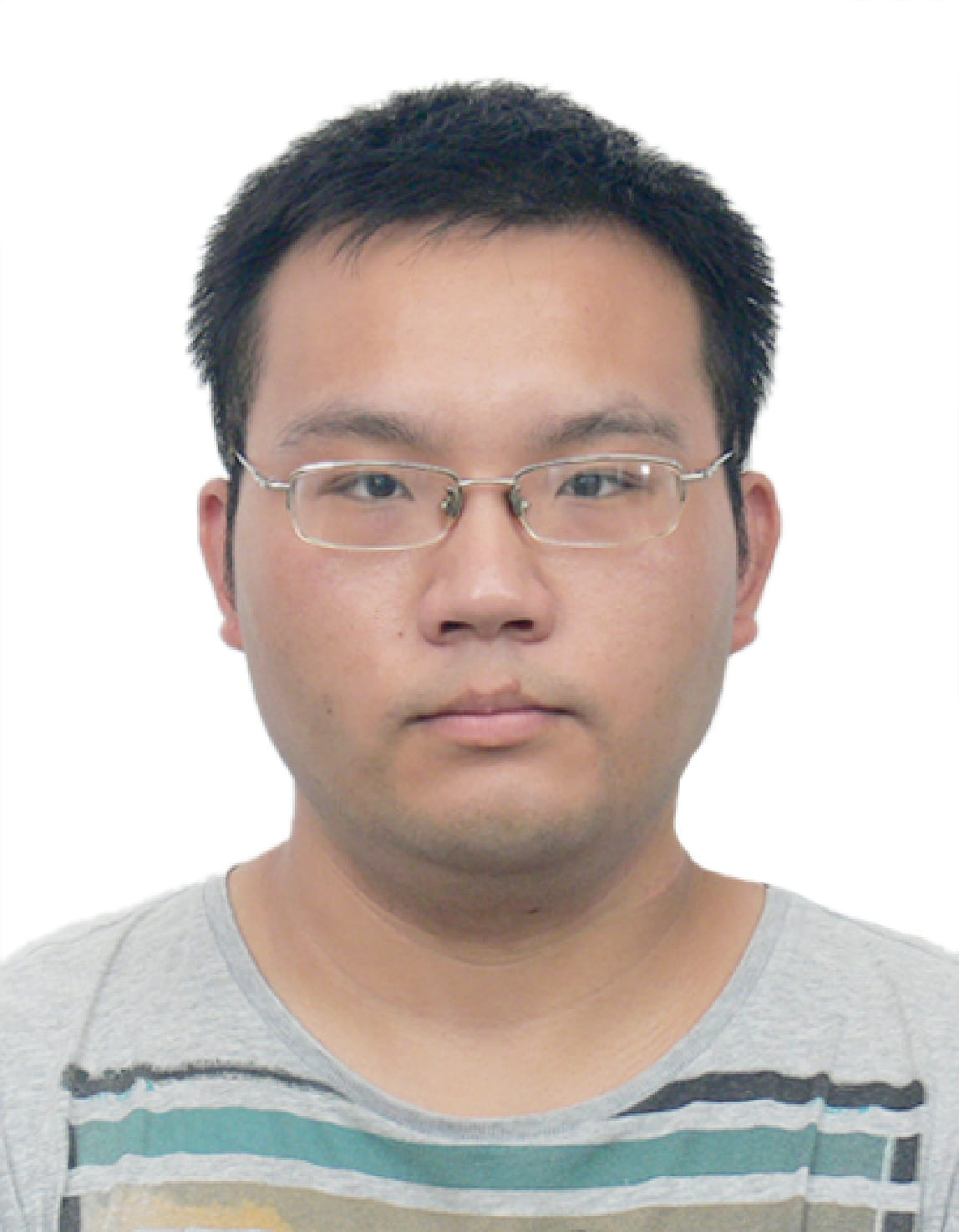}}] {Yuan Zong} (Member, IEEE) received the B.S. and M.S. degrees in electronics engineering from the Nanjing Normal University, Nanjing, China, in 2011 and 2014, respectively, and the Ph.D. degree in biomedical engineering from the Southeast University, Nanjing, China, in 2018. He is currently an Associate Professor with the Key Laboratory of Child Development and Learning Science of Ministry of Education, School of Biological Science and Medical Engineering, Southeast University. From 2016 to 2017, he was a Visiting Student with the Center for Machine Vision and Signal Analysis, University of Oulu, Oulu, Finland.
	
	He has authored or coauthored more than 30 articles in mainstream journals and conferences, such as the IEEE T-IP, T-CYB, T-AFFC, AAAI, IJCAI, and ACM MM. His research interests include affective computing, pattern recognition, computer vision, and speech signal processing.
\end{IEEEbiography}

\begin{IEEEbiography}[{\includegraphics[width=1in,height=1.25in,clip,keepaspectratio]{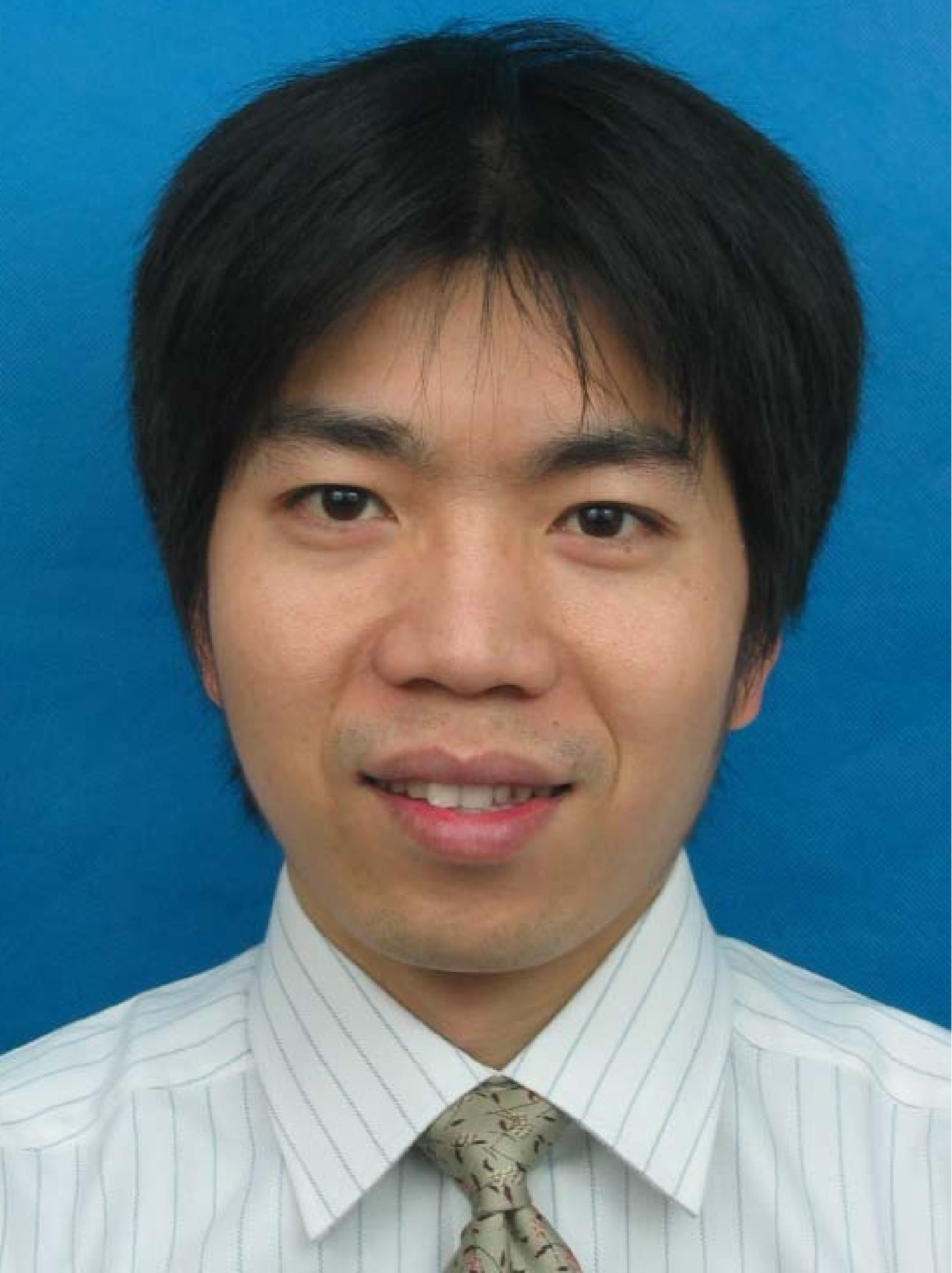}}] {Wenming Zheng} (Senior Member, IEEE) received the B.S. degree in computer science from Fuzhou University, Fuzhou, China, in 1997, the M.S. degree in computer science from Huaqiao University, Quanzhou, China, in 2001, and the Ph.D. degree in signal processing from Southeast University, Nanjing, China, in 2004. Since 2004, he has been with
	the Research Center for Learning Science, Southeast University, where he is currently a Professor with the School of Biological Science and Medical Engineering, and the Key Laboratory of Child Development and Learning Science of Ministry of Education. He has been elected as IET Fellow since 2022.
	His research interests include affective computing, pattern recognition, machine learning and computer vision. He is currently an Associated Editor for
	the IEEE TRANSACTIONS ON AFFECTIVE COMPUTING and the Editorial Board Member of \textit{The Visual Computer}.
\end{IEEEbiography}

\begin{IEEEbiography}[{\includegraphics[width=1in,height=1.25in,clip,keepaspectratio]{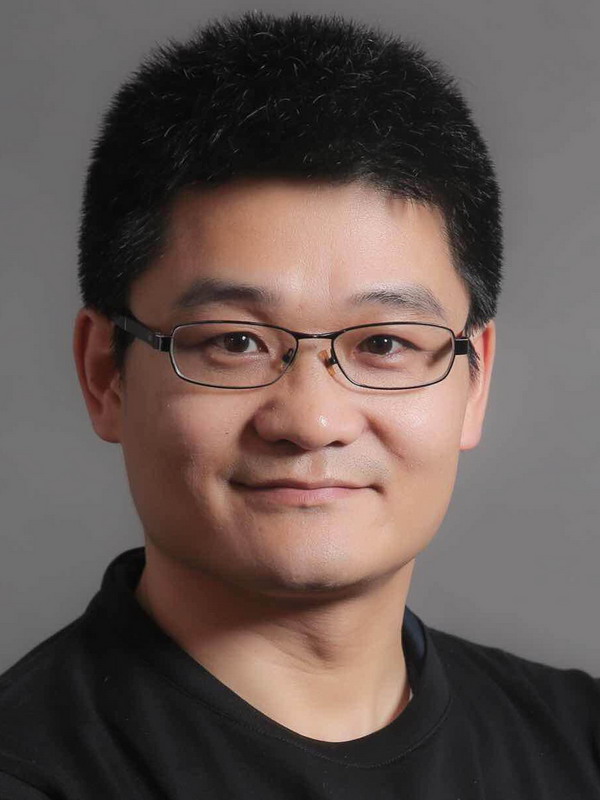}}]
	{Shiguang Shan} received M.S. degree in computer science from the Harbin Institute of Technology, Harbin, China, in 1999, and Ph.D. degree in computer science from the Institute of Computing Technology (ICT), Chinese Academy of Sciences (CAS), Beijing, China, in 2004. He joined ICT, CAS in 2002 and has been a Professor since 2010. He is now the deputy director of the Key Lab of Intelligent Information Processing of CAS. His research interests cover computer vision, pattern recognition, and machine learning. He has published more than 300 papers in refereed journals and proceedings in the areas of computer vision and pattern recognition. He has served as Area Chair for many international conferences including CVPR, ICCV, IJCAI, AAAI, ICPR, ACCV, FG, etc.. He is Associate Editors of several international journals including IEEE Trans. on Image Processing, Computer Vision and Image Understanding, Neurocomputing, and Pattern Recognition Letters. He is a recipient of the China's State Natural Science Award in 2015, and the China’s State S\&T Progress Award in 2005 for his research work. 
\end{IEEEbiography}

\begin{IEEEbiography}
[{\includegraphics[width=1in,height=1.25in,clip,keepaspectratio]{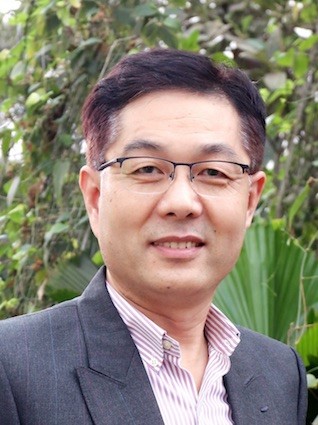}}]{Cuntai Guan}(Fellow, IEEE) received his Ph.D. degree from Southeast University, China, in 1993. He is a President’s Chair Professor in the School of Computer Science and Engineering, Nanyang Technological University, Singapore. He is the Director of the Artificial Intelligence Research Institute, Director of the Centre for Brain-Computing Research, and Co-Director of S-Lab for Advanced Intelligence. His research interests include brain-computer interfaces, machine learning, medical signal and image processing, artificial intelligence, and neural and cognitive rehabilitation. He is a recipient of the Annual BCI Research Award (first prize), King Salman Award for Disability Research, IES Prestigious Engineering Achievement Award, Achiever of the Year (Research) Award, and Finalist of President Technology Award. He is also an elected Fellow of the US National Academy of Inventors (NAI), the Academy of Engineering Singapore (SAEng), and the American Institute for Medical and Biological Engineering (AIMBE).
\end{IEEEbiography}
	
\end{document}